\documentclass[format=acmsmall,review=false,screen=true,natbib=true,nonacm=true]{acmart}
\usepackage{subcaption}
\usepackage{tikz}
\usetikzlibrary{arrows,calc}
\usepackage{multirow}
\usepackage{booktabs}
\begin{document}

\title{A Survey of Unsupervised Deep Domain Adaptation}

\author{Garrett Wilson}
\orcid{0000-0002-6760-754X}
\email{garrett.wilson@wsu.edu}
\author{Diane J. Cook}
\orcid{0000-0002-4441-7508}
\email{djcook@wsu.edu}
\affiliation{%
  \institution{Washington State University}
  \streetaddress{School of Electrical Engineering and Computer Science}
  \city{Pullman}
  \state{WA}
  \postcode{99164}
  \country{USA}}

\begin{abstract}
Deep learning has produced state-of-the-art results for a variety of tasks. While such approaches for supervised learning have performed well, they assume that training and testing data are drawn from the same distribution, which may not always be the case. As a complement to this challenge, single-source unsupervised domain adaptation can handle situations where a network is trained on labeled data from a source domain and unlabeled data from a related but different target domain with the goal of performing well at test-time on the target domain. Many single-source and typically homogeneous unsupervised deep domain adaptation approaches have thus been developed, combining the powerful, hierarchical representations from deep learning with domain adaptation to reduce reliance on potentially-costly target data labels. This survey will compare these approaches by examining alternative methods, the unique and common elements, results, and theoretical insights. We follow this with a look at application areas and open research directions.
\end{abstract}

\maketitle

\section{Introduction}
Supervised learning is arguably the most prevalent type of machine learning and has enjoyed much success across diverse application areas. However, many supervised learning methods make a common assumption: the training and testing data are drawn from the same distribution. When this constraint is violated, a classifier trained on the \textit{source domain} will likely experience a drop in performance when tested on the \textit{target domain} due to the differences between domains \cite{patel2015ieee}. Single-source \textit{domain adaptation} refers to the goal of learning a concept from labeled data in a source domain that performs well on a different but related target domain \cite{pan2010tkde,goodfellow2016deep,ganin2016jmlr}. \textit{Unsupervised} domain adaptation specifically addresses the situation where there are labeled source data and only unlabeled target data available for use during training \cite{long2015icml,ganin2016jmlr}.

Because of its ability to adapt labeled data for use in a new application, domain adaptation can reduce the need for costly labeled data in the target domain. As an example, consider the problem of semantically segmenting images. Each real image in the Cityscapes dataset required approximately 1.5 hours to annotate for semantic segmentation \cite{cordts2016cvpr}. In this case, human annotation time could be spared by training an image semantic segmentation model on synthetic street view images (the source domain) since these can be cheaply generated, then adapting and testing for real street view images (the target domain, here the Cityscapes dataset).

An undeniable trend in machine learning is the increased usage of deep neural networks. Deep networks have produced many state-of-the-art results for a variety of machine learning tasks \cite{ganin2016jmlr,goodfellow2016deep} such as image classification, speech recognition, machine translation, and image generation \cite{goodfellow2016deep,goodfellow2016survey}. When trained on large amounts of data, these many-layer neural networks can learn powerful, hierarchical representations \cite{sun2016,long2015icml,goodfellow2016deep,patel2015ieee} and can be highly scalable \cite{ghifary2016}. At the same time, these networks can also experience performance drops due to domain shifts \cite{sun2016,ganin2015icml}. Thus, much research has gone into adapting such networks from large labeled datasets to domains where little (or possibly no) labeled training data are available (for a list, see \cite{dapapers}). These single-source and typically homogeneous unsupervised deep domain adaptation approaches, which combine the benefit of deep learning with the very practical use of domain adaptation to remove the reliance on potentially costly target data labels, will be the focus of this survey.

A number of surveys have been created on the topic of domain adaptation \cite{margolis2011literature,beijbom2012domain,patel2015ieee,csurka2017domain,csurka2017comprehensive,wang2018deepsurvey,zhao2018unsupervised,kouw2018introduction,bungum2011survey,chu2018survey,sun2015multisourcesurvey,kouw2019review} and more generally transfer learning \cite{pan2010tkde,lu2015tlsurvey,shao2015tlsurvey,weiss2016survey,zhang2017transfer,tan2018survey,cook2013survey,taylor2009transfer,lazaric2012rlsurvey}, of which domain adaptation can be viewed as a special case \cite{patel2015ieee}. Previous domain adaptation surveys lack depth of coverage and comparison of unsupervised deep domain adaptation approaches. In some cases, prior surveys do not discuss domain mapping \cite{kouw2018introduction,csurka2017domain,csurka2017comprehensive}, normalization statistic-based \cite{kouw2018introduction,zhao2018unsupervised,csurka2017domain,csurka2017comprehensive}, or ensemble-based \cite{kouw2018introduction,zhao2018unsupervised,csurka2017domain,csurka2017comprehensive,wang2018deepsurvey} methods. In other cases, they do not survey deep learning approaches \cite{margolis2011literature,beijbom2012domain,patel2015ieee,kouw2019review}. Still others are application-centric, focusing on a single use case such as machine translation \cite{bungum2011survey,chu2018survey}. One earlier survey focuses on the multi-source scenario \cite{sun2015multisourcesurvey}, while we focus on the more prevalent single-source scenario. Transfer learning is a broader topic to cover, thus surveys provide minimal coverage and comparison of the deep learning methods that have been designed for unsupervised domain adaptation \cite{pan2010tkde,lu2015tlsurvey,shao2015tlsurvey,weiss2016survey,zhang2017transfer,tan2018survey}, or they focus on tasks such as activity recognition \cite{cook2013survey} or reinforcement learning  \cite{taylor2009transfer,lazaric2012rlsurvey}. The goal of this survey is to discuss, highlight unique components, and compare approaches to single-source homogeneous unsupervised deep domain adaptation.

We first provide background on where domain adaptation fits into the more general problem of transfer learning. We follow this with an overview of generative adversarial networks (GANs) to provide background for the increasingly widespread use of adversarial techniques in domain adaptation. Next, we investigate the various domain adaptation methods, the components of those methods, and the results. Then, we overview domain adaptation theory and discuss what we can learn from the theoretical results. Finally, we look at application areas and identify future research directions for domain adaptation.

\section{Background}

\subsection{Transfer Learning} \label{transferlearning}
The focus of this survey is domain adaptation. Because domain adaptation can be viewed as a special case of transfer learning \cite{patel2015ieee}, we first review transfer learning to highlight the role of domain adaptation within this topic. Transfer learning is defined as the learning scenario where a model is trained on a source domain or task and evaluated on a different but related target domain or task, where either the tasks or domains (or both) differ \cite{pan2010tkde,dredze2010multi,weiss2016survey,goodfellow2016deep}. For instance, we may wish to learn a model on a handwritten digit dataset (e.g., MNIST \cite{lecun1998mnist}) with the goal of using it to recognize house numbers (e.g., SVHN \cite{netzer2011reading}). Or, we may wish to learn a model on a synthetic, cheap-to-generate traffic sign dataset \cite{moiseev2013evaluation} with the goal of using it to classify real traffic signs (e.g., GTSRB \cite{Stallkamp-IJCNN-2011}). In these examples, the source dataset used to train the model is related but different from the target dataset used to test the model -- both are digits and signs respectively, but each dataset looks significantly different. When the source and target differ but are related, then transfer learning can be applied to obtain higher accuracy on the target data.

\subsubsection{Categorizing Methods}
In a transfer learning survey paper, Pan et al. \cite{pan2010tkde} defined two terms to help classify various transfer learning techniques: ``domain'' and ``task.'' A domain consists of a feature space and a marginal probability distribution (i.e., the features of the data and the distribution of those features in the dataset). A task consists of a label space and an objective predictive function (i.e., the set of labels and a predictive function that is learned from the training data). Thus, a transfer learning problem might be either transferring knowledge from a source domain to a different target domain or transferring knowledge from a source task to a different target task (or a combination of the two) \cite{pan2010tkde,dredze2010multi,weiss2016survey}.

By this definition, a change in domain may result from either a change in feature space or a change in the marginal probability distribution. When classifying documents using text mining, a change in the feature space may result from a change in language (e.g., English to Spanish), whereas a change in the marginal probability distribution may result from a change in document topics (e.g., computer science to English literature) \cite{pan2010tkde}. Similarly, a change in task may result from either a change in the label space or a change in the objective predictive function. In the case of document classification, a change in the label space may result from a change in the number of classes (e.g., from a set of 10 topic labels to a set of 100 topic labels). Similarly, a change in the objective predictive function may result from a substantial change in the distribution of the labels (e.g., the source domain has 100 instances of class A and 10,000 of class B, whereas the target has 10,000 instances of A and 100 of B) \cite{pan2010tkde}.

To classify transfer learning algorithms based on whether the task or domain differs between source and target, Pan et al. \cite{pan2010tkde} introduced three terms: ``inductive'', ``transductive'', and ``unsupervised'' transfer learning. In inductive transfer learning, the target and source tasks are different, the domains may or may not differ, and some labeled target data are required. In transductive transfer learning, the tasks remain the same while the domains are different, and both labeled source data and unlabeled target data are required. Finally, in unsupervised transfer learning, the tasks differ as in the inductive case, but there is no requirement of labeled data in either the source domain or the target domain.

\subsubsection{Domain Adaptation}
One popular type of transfer learning is \textit{domain adaptation}, which will be the focus of our survey. Domain adaptation is a type of transductive transfer learning. Here, the target task remains the same as the source, but the domain differs \cite{patel2015ieee,pan2010tkde,daume2006jail}. Homogeneous domain adaptation is the case where the domain feature space also remains the same, and heterogeneous domain adaptation is the case where the feature spaces differ \cite{patel2015ieee}.

In addition to the previous terminology, machine learning techniques are often categorized based on whether or not labeled training data are available. Supervised learning assumes labeled data are available, semi-supervised learning uses both labeled data and unlabeled data, and unsupervised learning uses only unlabeled data. However, domain adaptation assumes data comes from both a source domain and a target domain. Thus, prepending one of these three terms to ``domain adaptation'' is ambiguous since it may refer to labeled data being available in the source or target domains.

Authors apply these terms in various ways to domain adaptation \cite{jiang2008domain,pan2010tkde,saito2017icml,daume2007acl,weiss2016survey}. In this paper, we will refer to ``unsupervised'' domain adaptation as the case in which both labeled source data and unlabeled target data are available, ``semi-supervised'' domain adaptation as the case in which labeled source data in addition to some labeled target data are available, and ``supervised'' domain adaptation as the case in which both labeled source and target data are available \cite{beijbom2012domain}. The distinction between these categories describes the target domain, but only describe situations in which labeled data are available for the source domain. These definitions are commonly used in the methods surveyed in this paper as well as others   \cite{sun2016,saito2017icml,long2015icml,ganin2016jmlr,ghifary2016,carlucci2017autodial}.

\subsubsection{Related Problems}
Multi-domain learning \cite{dredze2010multi,joshi2012multi} and multi-task learning \cite{caruana1997multitask} are related to transfer learning and domain adaptation. In contrast to transfer learning, the goal of these learning approaches is obtaining high performance on all specified domains (or tasks) rather than just on a single target domain (or task) \cite{pan2010tkde,yang2015iclr}. For example, often it is assumed that the training data are drawn in an independent and identically distributed (i.i.d.) fashion, which may not be the case \cite{joshi2012multi}. One such example is the task of developing a spam filter for users who disagree on what is considered spam. If all the users' data are combined, the training data will be drawn from multiple domains. While each individual domain may be i.i.d., the aggregated dataset may not be. If the data are split by user, then there may be too little data to learn a model for each user. Multi-domain learning can take advantage of the entire dataset to learn individual user preferences \cite{dredze2010multi,joshi2012multi}. Some researchers have developed adversarial strategies to tackle this multi-domain learning challenge \cite{sebag2019multi,hassan2018unsupervised}.

When working with multiple tasks, instead of training models separately for different tasks (e.g., one model for detecting shapes in an image and one model for detecting text in an image), multi-task learning will learn these separate but related tasks simultaneously so that they can mutually benefit from the training data of other tasks through a (partially) shared representation \cite{caruana1997multitask}. If there are both multiple tasks and domains, then these approaches can be combined into multi-domain multi-task learning, as is described by Yang et al. \cite{yang2015iclr}.

Another related problem is domain generalization, in which a model is trained on multiple source domains with labeled data and then tested on a separate target domain that was not seen during training \cite{muandet2013domain}. This contrasts with domain adaptation where target examples (possibly unlabeled) are available during training. Some approaches related to those surveyed in this paper have been designed to address this situation. Examples include an adversarial method introduced by Zhao et al. \cite{zhao2017icml} and an autoencoder approach by Ghifary et al. \cite{ghifary2015iccv} discussed in Section~\ref{domainGeneralization}.

\subsection{Generative Adversarial Networks} \label{gan}
Many deep domain adaptation methods that we will discuss in the next section incorporate adversarial training. We use the term \textit{adversarial training} broadly to refer to any method that utilizes an adversary or an adversarial process during training. Before other adversarial methods were developed, the term was narrowly applied to training designed to improve the robustness of a model by utilizing adversarial examples, e.g. image inputs with small worst-case perturbations that lead to misclassification \cite{szegedy2013intriguing,goodfellow2014explaining}. Subsequently, other techniques have arisen that also utilize an adversary during training, including generative-adversarial training of generative adversarial networks (GANs) \cite{goodfellow2014nips} and domain-adversarial training of domain adversarial neural networks (DANN) \cite{ganin2016jmlr}, both of which have been used for domain adaptation. To provide background for the domain adaptation methods utilizing these techniques, we will first discuss GANs and later when discussing DANN note the differences.

In recent years there has been a large and growing interest in GANs. Pitting two well-matched neural networks against each other (hence ``adversarial''), playing the roles of a data discriminator and a data generator, the pair is able to refine each player's abilities in order to perform functions such as synthetic data generation. Goodfellow et al. \cite{goodfellow2014nips} proposed this technique in 2014. Since that time, hundreds of papers have been published on the topic \cite{ganzoo,adversarialnetspapers}. GANs have traditionally been applied to synthetic image generation, but recently researchers have been exploring other novel use cases such as domain adaptation.

\begin{figure}
    \centering
    \includegraphics[width=0.75\linewidth]{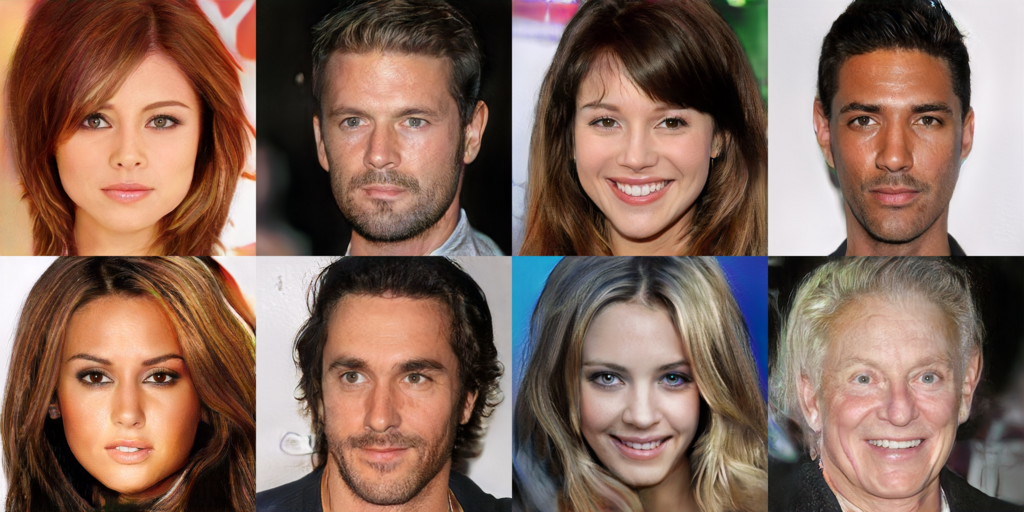}
    \caption{Realistic but entirely synthetic images of human faces generated by a GAN trained on the CelebA-HQ dataset \cite{karras2018progressive}.}
    \label{fig:Karras}
\end{figure}

GANs are a type of deep generative model \cite{goodfellow2014nips}. For synthetic image generation, a training dataset of images must be available. Popular datasets include human faces (CelebA \cite{Liu_2015_ICCV}), handwritten digits (MNIST \cite{lecun1998mnist}), bedrooms (LSUN \cite{Yu2015LSUNCO}), and sets of other objects (CIFAR-10 \cite{krizhevsky2009learning} and ImageNet \cite{5206848,ILSVRC15}). After training, the generative model will be able to generate synthetic images that resemble those in the training data. For example, a generator trained with CelebA will generate images of human faces that look realistic but are not images of real people, as shown in Figure~\ref{fig:Karras}. To learn to do this, GANs utilize two neural networks competing against each other \cite{goodfellow2014nips}. One network represents a generator. The generator accepts a noise vector as input, which contains random values drawn from some distribution such as normal or uniform. The goal of the generator network is to output a vector that is indistinguishable from the real training data. The other network represents a discriminator, which accepts as input either a real sample from the training data or a fake sample from the generator. The goal of the discriminator is to determine the probability that the input sample is real. During training, these two networks play a minimax game, where the generator tries to fool the discriminator and the discriminator attempts to not be fooled.

Using the notation from Goodfellow et al. \cite{goodfellow2014nips}, we define a value function $V(G,D)$ employed by the minimax game between the two networks:
\begin{align}
\min_G \max_D V(D,G) =
	\mathbb{E}_{x \sim p_\text{data}(x)} \left[ \log D(x) \right]
  + \mathbb{E}_{z \sim p_z(z)} \left[ \log(1 - D(G(z))) \right] \label{GANeq}
\end{align}

\begin{figure}
    \centering
    \tikzstyle{int}=[draw, minimum size=3em]
    \tikzstyle{init} = [pin edge={to-,thin,black}]

    \begin{tikzpicture}[node distance=2.25cm,auto,>=latex',scale=0.8,every node/.style={scale=0.8}]
        \node (a) {z};
        \node [int] (b) [right of=a] {G};
        \node (c) [right of=b] {\shortstack{\includegraphics[width=0.1\linewidth]{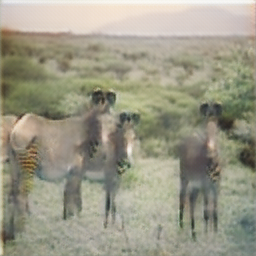} \\ \small{Fake Image}}};
        \node [int] (d) [right of=c] {D};
        \node (e) [right of=d] {\shortstack{``real''\\or\\``fake''}};
        \path[->] (a) edge node {} (b);
        \path[->] (b) edge node {} (c);
        \path[->] (c) edge node {} (d);
        \path[->] (d) edge node {} (e);

        \node (c2) [above of=c] {\shortstack{\includegraphics[width=0.1\linewidth]{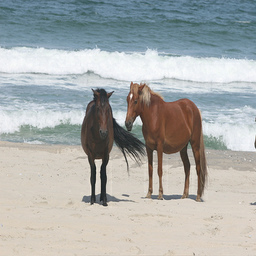} \\ \small{Real Image}}};
        \node [int] (d2) [right of=c2] {D};
        \node (e2) [right of=d2] {\shortstack{``real''\\or\\``fake''}};
        \path[->] (c2) edge node {} (d2);
        \path[->] (d2) edge node {} (e2);
        \path[dashed,-] (d) edge node {} (d2);
    \end{tikzpicture}
    \caption{Illustration of the GAN generator $G$ and discriminator $D$ networks. The dashed line between the $D$ networks indicates that they share weights (or are the same network). In the top row, a real image from the training data (horses $\leftrightarrow$ zebras dataset by Zhu et al. \cite{zhu2017iccv}) is fed to the discriminator, and the goal of $D$ is to make $D(x)=1$ (correctly classify as real). In the bottom row, a fake image from the generator is fed to the discriminator, and the goal of $D$ is to make $D(G(z))=0$ (correctly classify as fake), which competes with the goal of $G$ to make $D(G(z))=1$ (misclassify as real).}
    \label{fig:GAN}
\end{figure}
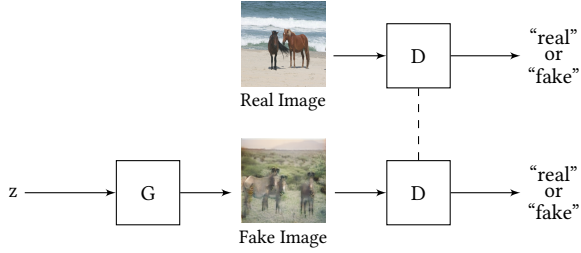

Here, $x \sim p_{data}(x)$ draws a sample from the real data distribution, $z \sim p_z(z)$ draws a sample from the input noise, $D(x;\theta_d)$ is the discriminator, and $G(z;\theta_g)$ is the generator. As shown in the equation, the goal is to find the parameters $\theta_d$ that maximize the log probability of correctly discriminating between real ($x$) and fake ($G(z)$) samples while at the same time finding the parameters $\theta_g$ that minimize the log probability of $1-D(G(z))$. The term $D(G(z))$ represents the probability that generated data $G(z)$ is real. If the discriminator correctly classifies a fake input then $D(G(z))=0$. Equation \ref{GANeq} minimizes the quantity $1-D(G(z))$. This occurs when $D(G(z))=1$, or when the discriminator misclassifies the generator's output as a real sample. Thus the discriminator's mission is to learn to correctly classify the input as real or fake while the generator tries to fool the discriminator into thinking that its generated output is real. This process is illustrated in Figure \ref{fig:GAN}.

\subsubsection{Training} \label{training}
In recent years there have been impressive results from GANs. At the same time, this research faces some challenges since training a GAN can encounter problems such as difficulty converging \cite{goodfellow2016survey,arora2017icml}, mode collapse where the generator only learns to generate realistic samples for a few specialized modes of the data distribution \cite{goodfellow2016survey}, and vanishing gradients \cite{goodfellow2014nips}. Many methods have been proposed to resolve these training challenges using a variety of tricks \cite{goodfellow2014nips,salimans2016nips,szegedy2016cvpr,pmlr-v70-odena17a,shrivastava2017cvpr,heusel2017nips}, network architecture choices \cite{radford2015,salimans2016nips,karras2018progressive}, objective modifications \cite{zhao2016iclr,berthelot2017,metz2016,mao2017least,nowozin2016nips,arjovsky2017icml,gulrajani2017nips,kodali2017,fedus2017many,jolicoeur2018relativistic,miyato2018spectral,odena2018generator,nguyen2017cvpr}, mixtures or ensembles \cite{ghosh2018multi,hoang2018mgan,park2018megan,khayatkhoei2018disconnected,mordido2018dropout,durugkar2017generative,arora2017icml,zhang2018generative,tolstikhin2017adagan}, maximum mean discrepancy (MMD) \cite{dziugaite2015training,li2015generative,sutherland2016generative,li2017mmd,binkowski2018demystifying}, making a connection to reinforcement learning \cite{finn2016,pfau2016}, or a combination of these modifications \cite{miyato2018cgans,heusel2017nips,zhang2018self}. For an in-depth discussion of these techniques, there are a number of survey papers directed at GAN variants that include a discussion of training challenges and work \cite{hong2017generative,manisha2018,hitawala2018}. These techniques can be employed in the domain adaptation methods that utilize GANs \cite{liu2016nips,mao2018unpaired,shrivastava2017cvpr,bousmalis2017cvpr,hoffman2018icml,bousmalis2018roboticgrasping,sankaranarayanan2018cvpr,wang2018domain,wei2018generative,choi2018cvpr}. While these training stability methods could similarly be applied to other adversarial domain adaptation approaches, they are not typically needed for the non-GAN methods surveyed here.

\subsubsection{Evaluation} \label{evaluation}
Once successfully trained, a GAN model can be difficult to evaluate and compare with other models. Multiple approaches and measures have been introduced to evaluate GAN performance. Often researchers have evaluated their models through visual inspection \cite{pmlr-v80-santurkar18a} such as performing user studies where participants mark which images they think look more realistic \cite{salimans2016nips}. However, ideally a more automated metric could be found. Past generative models were evaluated by computing log-likelihood \cite{theis2016iclr}, but this is not necessarily tractable in GANs \cite{goodfellow2016survey}. A proxy for log-likelihood is a Parzen window estimate, which was used for early GAN evaluation \cite{theis2016iclr,goodfellow2014nips,makhzani2015,nowozin2016nips}, but in high dimensions (such as images), this could be far from the actual log-likelihood and not even rank models correctly \cite{theis2016iclr,grover2017flow}. Thus, there has been much work proposing various evaluation methods for GANs: methods for detecting memorization \cite{goodfellow2014nips,makhzani2015,donahue2017iclr,theis2016iclr,radford2015,berthelot2017}, determining diversity \cite{arora2018,pmlr-v80-santurkar18a,pmlr-v70-odena17a,heusel2017nips}, measuring realism \cite{salimans2016nips,heusel2017nips,liu2018,binkowski2018demystifying}, and approximating log-likelihood \cite{wu2017iclr}. Xu et al. \cite{xu2018empirical} and Borji \cite{borji2018} survey and compare many of these GAN evaluation methods.

These techniques can be used for evaluating domain adaptation methods used for image translation (a form of image generation but conditioned on an input image) from one domain to another \cite{yoo2016pixel,zhu2017iccv,yi2017iccv,choi2018cvpr,royer2017xgan,benaim2017nips}. However, many domain adaptation methods (even those that are adversarial such as those using GANs) are not used for generation but rather for tasks with more easily-defined loss functions, making these techniques largely not needed for adversarial domain adaptation methods. For example, accuracy \cite{liu2016nips,ganin2016jmlr,tzeng2017cvpr,bousmalis2016nips,bousmalis2017cvpr,hoffman2018icml,choi2018cvpr,benaim2017nips,fu2018geometry} or AUC scores \cite{purushotham2017variational} can be used to evaluate classification, intersection over union or pixel accuracy can be used to evaluate image segmentation \cite{hoffman2018icml,benaim2017nips,fu2018geometry,li2018semantic,perone2018unsupervised}, and absolute difference can be used to evaluate regression \cite{shrivastava2017cvpr}.

\section{Methods} \label{methods}
In recent years, numerous new unsupervised domain adaptation methods have been proposed, with a growing emphasis on neural network-based approaches. Distinct lines of research have emerged. These include aligning the source domain and target domain distributions, mapping between domains, separating normalization statistics, designing ensemble-based methods, or focusing on making the model target discriminative by moving the decision boundary into regions of lower data density. In addition, others have explored combinations of these approaches. We will describe each of these categories together with recent methods that fall into these categories.

In this survey, we will focus on homogeneous domain adaptation consisting of one source and one target domain, as is most commonly studied. Another case is multi-source domain adaptation, where there are multiple source domains but still only one target domain. Sun et al. \cite{sun2015multisourcesurvey} survey multi-source domain adaptation, and since then a number of other methods \cite{guo2018multi,hoffman2018multisource,zhao2018multisource,peng2018moment,carlucci2018agnostic,mancini2018cvpr,xie2017nips,xu2018deepcocktail,redko2018optimal} have been developed for this case. It is also possible to perform multi-target domain adaptation \cite{gholami2018unsupervised}, though this case is even more rarely studied. Similarly, we focus on homogeneous domain adaptation due to its prevalence, though some heterogeneous methods have been developed \cite{yao2020dda,zhou2019heterogeneous,tsai2016cvpr,duan2012augmented,wang2011heterogeneous,li2019heterogeneous}.

\subsection{Domain-Invariant Feature Learning} \label{domainInvariance}
Most recent domain adaptation methods align source and target domains by creating a domain-invariant feature representation, typically in the form of a feature extractor neural network. A feature representation is domain-invariant if the features follow the same distribution regardless of whether the input data are from the source or target domain \cite{zhao2019learning}. If a classifier can be trained to perform well on the source data using domain-invariant features, then the classifier may generalize well to the target domain since the features of the target data match those on which the classifier was trained. However, these methods assume that such a feature representation exists and the marginal label distributions do not differ significantly (Section~\ref{theory}).

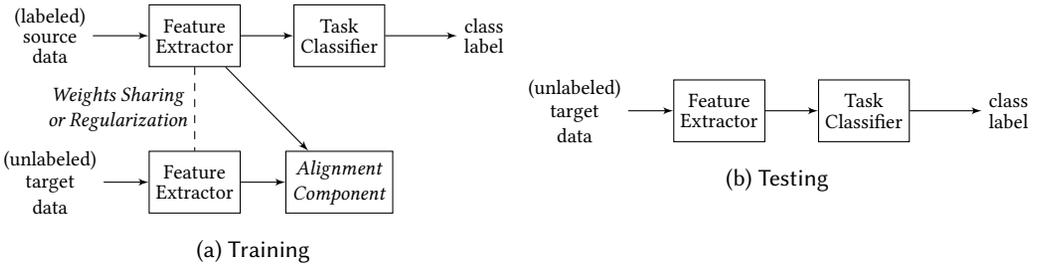
\begin{figure}
    \tikzstyle{int}=[draw, minimum size=3em]
    \tikzstyle{init} = [pin edge={to-,thin,black}]

    \begin{subfigure}{0.5\textwidth}
        \centering
        \begin{tikzpicture}[node distance=2.5cm,auto,>=latex',scale=0.77,every node/.style={scale=0.77}]
            \node (a) {\shortstack{(labeled) \\ source \\ data}};
            \node (b) [below of=a] {\shortstack{(unlabeled) \\ target \\ data}};
            \node [int] (c) [right of=a] {\shortstack{Feature \\ Extractor}};
            \node [int] (d) [right of=b] {\shortstack{Feature \\ Extractor}};
            \node [int] (e) [right of=c] {\shortstack{Task \\ Classifier}};
            \node [int] (f) [right of=d] {\shortstack{\textit{Alignment} \\ \textit{Component}}};
            \node (g) [right of=e] {\shortstack{class \\ label }};
            \path[->] (a) edge node {} (c);
            \path[->] (b) edge node {} (d);
            \path[->] (c) edge node {} (e);
            \path[->] (c) edge node {} (f);
            \path[->] (d) edge node {} (f);
            \path[->] (e) edge node {} (g);
            \draw[dashed,-] (c) -- (d) node [midway, left] (textNode1) {\shortstack{\textit{Weights Sharing} \\ \textit{or Regularization}}};
        \end{tikzpicture}
        \caption{Training}
    \end{subfigure}
    \begin{subfigure}{0.49\textwidth}
        \centering
        \begin{tikzpicture}[node distance=2.5cm,auto,>=latex',scale=0.77,every node/.style={scale=0.77}]
            \node (a) {\shortstack{(unlabeled) \\ target \\ data}};
            \node [int] (b) [right of=a] {\shortstack{Feature \\ Extractor}};
            \node [int] (c) [right of=b] {\shortstack{Task \\ Classifier}};
            \node (d) [right of=c] {\shortstack{class \\ label }};
            \path[->] (a) edge node {} (b);
            \path[->] (b) edge node {} (c);
            \path[->] (c) edge node {} (d);
        \end{tikzpicture}
        \caption{Testing}
    \end{subfigure}
    \caption{General network setup for domain adaptation methods learning domain-invariant features. (a) Methods differ in regard to how the domains are aligned during training (the Alignment Component) and whether the feature extractors used on each domain share none, some, or all of the weights between domains. (b) The target data are fed to the domain-invariant feature extractor and then to the task classifier.}
    \label{fig:alignment}
\end{figure}

The general training and testing setup of these methods is illustrated in Figure~\ref{fig:alignment}. Methods differ in how they align the domains (the Alignment Component in the figure). Some minimize divergence, some perform reconstruction, and some employ adversarial training. In addition, they differ in weight sharing choices, which will be discussed in Section~\ref{weightSharing}. We discuss the various alignment methods below.

\subsubsection{Divergence}
One method of aligning distributions is through minimizing a divergence that measures the distance between the distributions. Choices for the divergence measure include maximum mean discrepancy, correlation alignment, contrastive domain discrepancy, the Wasserstein metric, and a graph matching loss.

Maximum mean discrepancy (MMD) \cite{gretton2007kernel,gretton2012kernel} is a two-sample statistical test of the hypothesis that two distributions are equal based on observed samples from the two distributions. The test is computed from the difference between the mean values of a smooth function on the two domains' samples. If the means are different, then the samples are likely not from the same distribution. The smooth functions chosen for MMD are unit balls in characteristic reproducing kernel Hilbert spaces (RKHS) since it can be proven that the population MMD is zero if and only if the two distributions are equal \cite{gretton2012kernel}.

To use MMD for domain adaptation, the alignment component can be another classifier similar to the task classifier. MMD can then be computed and minimized between the outputs of these classifiers' corresponding layers (a slightly different setup than that in Figure~\ref{fig:alignment}). Rozantsev et al. \cite{rozantsev2018ieee} employ MMD, Long et al. \cite{long2015icml} investigate a multiple kernel variant of MMD (MK-MMD), and later Long et al. \cite{long2015icml} develop a joint MMD (JMMD) method \cite{long2017jmmd}. Bousmalis et al. \cite{bousmalis2016nips} also tried MMD but found using an adversarial objective performed better in their experiments.

Correlation alignment (CORAL) \cite{sun2016aaai} is similar to MMD with a polynomial kernel, computed from the distance between second-order statistics (covariances) of the source and target features. For domain adaptation, the alignment component consists of computing the CORAL loss between the two feature extractors' outputs (in order to minimize the distance). A variety of distances have been used: Sun et al. \cite{sun2016} use a squared matrix Frobenius norm in Deep CORAL, Zhang et al. \cite{zhang2018mca} use a Euclidean distance in mapped correlation alignment (MCA), others have used log-Euclidean distances in LogCORAL \cite{wang2017iccv} and Log D-CORAL\cite{morerio2017correlation}, and Morerio et al. \cite{morerio2018minimalentropy} use geodesic distances. Zhang et al. \cite{zhang2018aligning} generalize correlation alignment to possibly infinite-dimensional covariance matrices in RKHS. Chen et al. \cite{chen2019homm} align statistics beyond the first and second orders.

Contrastive domain discrepancy (CCD) \cite{kang2019contrastive} is based on MMD but looks at the conditional distributions in order to incorporate label information (unlike CORAL or ordinary MMD). When minimizing CCD, intra-class discrepancy is minimized while inter-class margin is maximized. This has the problem of requiring target labels though, so Kang et al. \cite{kang2019contrastive} propose contrastive adaptation networks (CAN) that minimize cross-entropy loss on the labeled target data while alternating between estimating labels for target samples (via clustering) with adapting the feature extractor with the now-computable CCD (using the clusters). This approach outperforms the other methods on the Office dataset as shown in Table~\ref{comparePerformance2}.

A problem known as ``optimal transport'' was originally proposed for studying resource allocation such as finding an optimal way to move material from mines to factories \cite{monge1781memoire,redko2017theoretical}, but it can also be used to measure the distances between distributions. If the cost of moving each point is a norm (e.g., Euclidean), then the solution to a discrete optimal transport problem can be viewed as a distance: the Wasserstein distance \cite{damodaran2018deepjdot} (also known as the earth mover's distance). To align feature and label distributions with this distance, Courty et al. \cite{courty2017nips} propose joint distribution optimal transport (JDOT). To incorporate this into a neural network, Damodaran et al. \cite{damodaran2018deepjdot} propose DeepJDOT.

Another divergence measure arises from graph matching: the problem of finding an optimal correspondence between graphs \cite{yan2016surveygraphmatching}. A feature extractor's output on a batch of samples can be viewed as an undirected graph (in the form of an adjacency matrix), where similar samples in the batch are connected. Given the graph from a batch of source data fed through the feature extractor and similarly a graph from a batch of target data, then the cost of aligning these graphs can be used as a divergence, as proposed by Das et al. \cite{das2018graphmatching,das2018sampletosample,das2018hypergraph}.

\subsubsection{Reconstruction}
Rather than minimizing a divergence, Ghifary et al. \cite{ghifary2016} and Bousmalis et al. \cite{bousmalis2016nips} hypothesize that alignment can be accomplished by learning a representation that both classifies the labeled source domain data well and can be used to reconstruct either the target domain data (Ghifary et al.) or both the source and target domain data (Bousmalis et al.). The alignment component in these setups is a reconstruction network -- the opposite of the feature extractor network -- that takes the feature extractor output and recreates the feature extractor's input (in this case, an image). Ghifary et al. \cite{ghifary2016} propose deep reconstruction-classification networks (DRCN), using a pair-wise squared reconstruction loss. Bousmalis et al. \cite{bousmalis2016nips} propose domain separation networks (DSN), using a scale-invariant mean squared error reconstruction loss.

\subsubsection{Adversarial} \label{featureLevelAdaptation}
Several varieties of feature-level adversarial domain adaptation methods have been introduced in the literature. In most the alignment component consists of a domain classifier. In one paper this component is instead represented by a network learning an approximate Wasserstein distance, and in another paper the component is a GAN.

A domain classifier is a classifier that outputs whether the feature representation was generated from source or target data. Recall that GANs include a discriminator that tries to accurately predict whether a sample is from the real data distribution or from the generator. In other words, the discriminator differentiates between two distributions, one real and one fake. A discriminator could similarly be designed to differentiate two distributions which instead represent a source distribution and a target distribution, as is done with a domain classifier. Note though that an adversarial domain classifier is used for adaptation, whereas a GAN is used for data generation. The domain classifier is trained to correctly classify the domain (source or target). In this scenario, the feature extractor is trained such that the domain classifier is unable to classify from which domain the feature representation originated. This is a type of zero-sum two-player game \cite{zhao2019learning} as in a GAN (Section~\ref{gan}). Typically, these networks are adversarially trained by alternating between these two steps. The feature extractor can be trained to make the domain classifier perform poorly by negating the gradient from the domain classifier with a \textit{gradient reversal layer} \cite{ganin2015icml} when performing back propagation to update the feature extractor weights (e.g., in DANN \cite{ajakan2014domain,ganin2015icml,ganin2016jmlr} and VRADA \cite{purushotham2017variational}), maximally confusing the domain classifier (when it outputs a uniform distribution over binary labels \cite{tzeng2015iccv}), or inverting the labels (in ADDA \cite{tzeng2017cvpr}). Because data distributions are often multi-modal, results may be improved by conditioning the domain classifier on a multilinear map of the feature representation and the task classifier predictions, which takes into account the multi-modal nature of the distributions \cite{long2018nips}.

Shen et al. \cite{shen2018wasserstein} created WDGRL, a modification of DANN, by replacing the domain classifier with a network that learns an approximate Wasserstein distance. This distance is then minimized between source and target domains, which they found to yield an improvement. This method is similar to the divergence methods except here the divergence is learned with a network rather than computed based on statistics (e.g., using mean in MMD or covariance in CORAL). This method outperforms the other methods on the Amazon review dataset as shown in Table~\ref{compareSentimentPerformance}.

Sankaranarayanan et al. \cite{sankaranarayanan2018cvpr} propose Generate to Adapt that uses a GAN as the alignment component. The feature extractor output is both fed to a classifier trained to predict the label (if the input is from the source domain) and also to a GAN trained to generate source-like images (regardless of if the input is source or target). For training stability, they use an AC-GAN \cite{pmlr-v70-odena17a}. They note one downside of using a GAN for adaptation is that it requires a large training dataset, but a common strategy is to use a pretrained network on a large dataset such as ImageNet. Using this pretraining, even on small datasets (e.g., Office) where the generated images are poor, the network still learns adaptation satisfactorily. Sankaranarayanan et al. \cite{sankaranarayanan2018cvprsemantic} similarly develop a similar approach for semantic segmentation.

\subsection{Domain Mapping} \label{domainMapping} \label{pixelLevelAdaptation}
\begin{figure}
    \centering
    \includegraphics[width=0.4\linewidth]{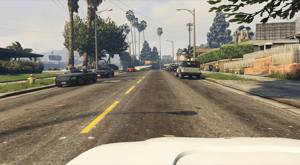}
    \includegraphics[width=0.4\linewidth]{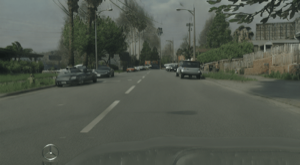}
    \caption{Synthetic vehicle driving image (left) adapted to look realistic (right) \cite{hoffman2018icml}.}
    \label{fig:driving}
\end{figure}

An alternative to creating a domain-invariant feature representation is mapping from one domain to another. The mapping is typically created adversarially and at the pixel level (i.e., pixel-level adversarial domain adaptation), but not always, as discussed at the end of this section. This mapping can be accomplished with a conditional GAN. The generator performs adaptation at the pixel level by translating a source input image to an image that closely resembles the target distribution. For example, the GAN could change from a synthetic vehicle driving image to one that looks realistic as shown in Figure~\ref{fig:driving} \cite{yoo2016pixel,zhu2017iccv,royer2017xgan,choi2018cvpr,hoffman2018icml}. A classifier can then be trained on the source data mapped to the target domain using the known source labels \cite{shrivastava2017cvpr} or jointly trained with the GAN \cite{bousmalis2017cvpr,hoffman2018icml}. We will first discuss how a conditional GAN works followed by the ways it can be employed for domain adaptation.

\subsubsection{Conditional GAN for Image-to-Image Translation}
The original formulation of a GAN was unconditional, where a GAN only accepted a noise vector as input. Conditional GANs, on the other hand, accept as input other information such as a class label, image, or other data \cite{goodfellow2014nips,gauthier2014conditional,mirza2014conditional,denton2015deep}. In the case of image generation, this means that a particular type of image to generate can be specified. One such example is to generate an image of a particular class within an image dataset such as ``cat'' rather than a random object from the dataset. Another example is conditioning on an input image such as in Figure~\ref{fig:driving}, mapping an input driving image from one domain (synthetic) to an output image in another domain (realistic). Other uses include: transferring style (e.g., make a photo look like a Van Gogh painting)  \cite{zhu2017iccv,yi2017iccv,kim2017discogan}, colorizing images \cite{isola2017cvpr}, generating satellite images from Google Maps data (or vice versa) \cite{isola2017cvpr,zhu2017iccv,yi2017iccv}, generating images of clothing from images of people wearing the clothing \cite{yoo2016pixel}, generating cartoon faces from real faces \cite{taigman2016dtn,royer2017xgan}, converting labels to photos (e.g., semantic segmentation output to a photo) \cite{isola2017cvpr,zhu2017iccv,yi2017iccv}, learning disentangled representations \cite{chen2016nips}, improving GAN training stability \cite{pmlr-v70-odena17a}, and domain adaptation, which will be discussed in Section~\ref{i2iforda}.

GANs conditioned on an input image can be used to perform image-to-image translation. These networks can be trained with varying levels of supervision: the dataset may contain corresponding images in the domains (supervised  \cite{yoo2016pixel,isola2017cvpr}), only a few corresponding images (semi-supervised  \cite{gan2017triangle}), or no corresponding images (unsupervised  \cite{zhu2017iccv,yi2017iccv,kim2017discogan}). A popular and general-purpose supervised method is pix2pix, developed by Isola et al. \cite{isola2017cvpr}. A commonly used unsupervised method is CycleGAN \cite{zhu2017iccv}, which is based on pix2pix, or methods similar to CycleGAN including DualGAN \cite{yi2017iccv} and DiscoGAN  \cite{kim2017discogan}.

Numerous modifications to these approaches have been proposed: one that is multi-modal is MUNIT, a multi-modal unsupervised image-to-image translator  \cite{huang2018multimodal}. By assuming a decomposition into style (domain-specific) and content (domain-invariant) codes, MUNIT can generate diverse outputs for a given input image (e.g., multiple possible output images corresponding to the same input image). A modification to CycleGAN explored by Li et al. \cite{li2018twin} uses separate batch normalization for each domain (an idea similar to AdaBN discussed in Section~\ref{normalizationStats}). Mejjati et al. \cite{mejjati2018nips} and Chen et al. \cite{chen2018eccv} improve results with attention, learning which areas of the images on which to focus. Shang et al. \cite{shang2017vigan} improve results by feeding the mapped images into a denoising autoencoder. While CycleGAN and similar approaches use two generators, one for each mapping direction, Benaim et al. \cite{benaim2017nips} developed a method for one-sided mapping that maintains distances between pairs of samples when mapped from the source to the target domain rather than (or in addition to) using a cycle consistency loss, and Fu et al. \cite{fu2018geometry} developed an alternative one-sided mapping using a geometric constraint (e.g., vertical flipping or 90 degree rotation). Royer et al. \cite{royer2017xgan} propose XGAN, a dual adversarial autoencoder capable of handling large domain shifts, where possibly an image in the source domain may correspond to multiple images in the target domain or vice versa. They tested mapping human faces to cartoon faces, which was a shift larger than CycleGAN could adequately handle. Choi et al. \cite{choi2018cvpr} propose StarGAN, a method for handling multiple domains with a single GAN. Approaches like CycleGAN need a separate generator (or two, one for each direction) for each pair of domains, which is not a scalable solution to many domains. StarGAN, on the other hand, only needs a single generator. This has the added benefit of allowing the generator to learn using all the available data rather than only the data in a specific pair of domains. During training they randomly pick a target domain at each iteration so the generator learns to generate images in all the domains. Anoosheh et al. \cite{anoosheh2018cvpr} propose an approach designed for the same purpose as StarGAN but using one generator per domain.

\subsubsection{Image-to-Image Translation for Domain Adaptation} \label{i2iforda}
While the above approaches map images from one domain to another without the explicit purpose of performing domain adaptation, they can also be used for domain adaptation. For example, the original CycleGAN paper was application agnostic, but others have experimented with applying CycleGAN to domain adaptation \cite{hoffman2018icml,benaim2017nips,fu2018geometry}. It is important to note though that these image-to-image translation approaches assume that the domain differences are primarily low-level \cite{bousmalis2017cvpr,bousmalis2018roboticgrasping,tzeng2017cvpr}.

\begin{figure}
    \tikzstyle{int}=[draw, minimum size=3em]
    \tikzstyle{init} = [pin edge={to-,thin,black}]

    \begin{subfigure}{\linewidth}
        \begin{minipage}{0.59\textwidth}
            \centering
            \begin{tikzpicture}[node distance=2.25cm,auto,>=latex',scale=0.77,every node/.style={scale=0.77}]
                \node (a) {\shortstack{(unlabeled) \\ source \\ data}};
                \node [int] (b) [right of=a] {\shortstack{Cond. \\ GAN}};
                \node (c) [right of=b] {\shortstack{(unlabeled) \\ target \\ data}};
                \path[->] (a) edge node {} (b);
                \path[->] (b) edge node {} (c);

                \node (d) [below of=a] {\shortstack{(labeled) \\ source \\ data}};
                \node [int] (e) [right of=d] {\shortstack{Cond. \\ GAN}};
                \node (f) [right of=e] {\shortstack{(labeled) \\ target \\ data}};
                \node [int] (g) [right of=f] {\shortstack{Target \\ Class.}};
                \node (h) [right of=g] {\shortstack{class \\ label}};
                \path[->] (d) edge node {} (e);
                \path[->] (e) edge node {} (f);
                \path[->] (f) edge node {} (g);
                \path[->] (g) edge node {} (h);
                \path[dashed,-] (b) edge node {} (e);
            \end{tikzpicture}
        \end{minipage}\hfill
        \begin{minipage}{0.39\textwidth}
            \centering
            \begin{tikzpicture}[node distance=2.25cm,auto,>=latex',scale=0.77,every node/.style={scale=0.77}]
                \node (a) {\shortstack{(unlabeled) \\ target \\ data}};
                \node [int] (b) [right of=a] {\shortstack{Target \\ Class.}};
                \node (c) [right of=b] {\shortstack{class \\ label}};
                \path[->] (a) edge node {} (b);
                \path[->] (b) edge node {} (c);
            \end{tikzpicture}
        \end{minipage}
        \caption{Method 1 (most common) -- training (left), testing (right)}
        \label{fig:mapping1}
        \vspace*{6mm}
    \end{subfigure}

    \begin{subfigure}{\linewidth}
        \begin{minipage}{0.39\textwidth}
            \centering
            \begin{tikzpicture}[node distance=2.25cm,auto,>=latex',scale=0.77,every node/.style={scale=0.77}]
                \node (a) {\shortstack{(unlabeled) \\ target \\ data}};
                \node [int] (b) [right of=a] {\shortstack{Cond. \\ GAN}};
                \node (c) [right of=b] {\shortstack{(unlabeled) \\ source \\ data}};
                \path[->] (a) edge node {} (b);
                \path[->] (b) edge node {} (c);

                \node (f) [below of=a] {\shortstack{(labeled) \\ source \\ data}};
                \node [int] (g) [right of=f] {\shortstack{Source \\ Class.}};
                \node (h) [right of=g] {\shortstack{class \\ label}};
                \path[->] (f) edge node {} (g);
                \path[->] (g) edge node {} (h);
            \end{tikzpicture}
        \end{minipage}\hfill
        \begin{minipage}{0.59\textwidth}
            \centering
            \begin{tikzpicture}[node distance=2.25cm,auto,>=latex',scale=0.77,every node/.style={scale=0.77}]
                \node (a) {\shortstack{(unlabeled) \\ target \\ data}};
                \node [int] (b) [right of=a] {\shortstack{Cond. \\ GAN}};
                \node (c) [right of=b] {\shortstack{(unlabeled) \\ source \\ data}};
                \node [int] (d) [right of=c] {\shortstack{Source \\ Class.}};
                \node (e) [right of=d] {\shortstack{class \\ label}};
                \path[->] (a) edge node {} (b);
                \path[->] (b) edge node {} (c);
                \path[->] (c) edge node {} (d);
                \path[->] (d) edge node {} (e);
            \end{tikzpicture}
        \end{minipage}
        \caption{Method 2 -- training (left), testing (right)}
        \label{fig:mapping2}
        \vspace*{6mm}
    \end{subfigure}

    \caption{Two possible configurations using image-to-image translation for domain adaptation. The conditional GAN and classifier can be trained separately or jointly. Method 1 is the most common. Method 2 is used by one paper. A combination of methods 1 and 2 is used in one paper. The dashed lines between networks indicate that they share weights (or are the same network). Note: this figure does not illustrate the many variants of the conditional GAN component, which often train a generator in each direction (one source to target and one target to source) and use additional losses such as cycle consistency.}
    \label{fig:mapping}
\end{figure}
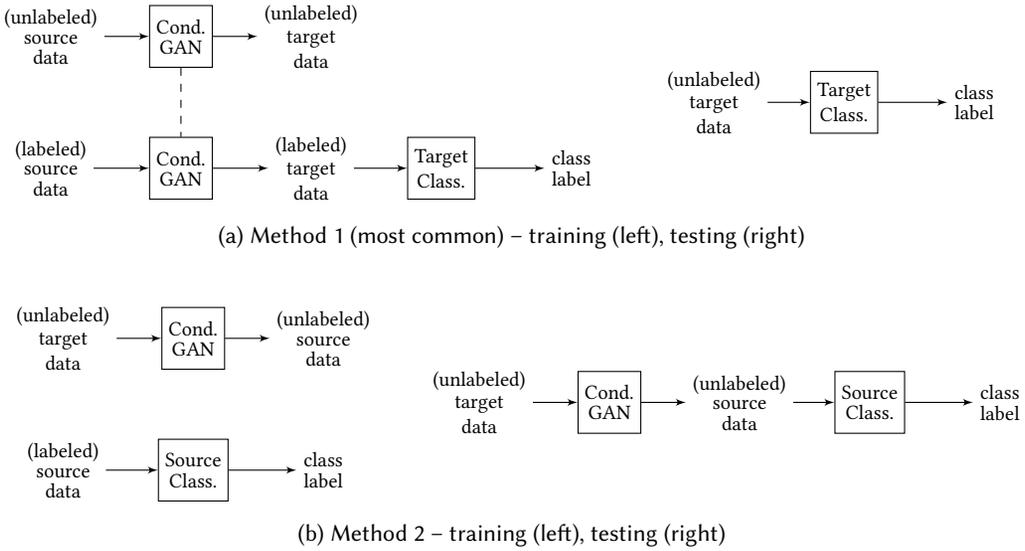

If unsupervised domain adaptation is performed for classification, adaptation can be accomplished by training an image-to-image translation GAN to map data from source to target, training a classifier on the mapped source images with known labels, and then subsequently testing by feeding unlabeled target through this target-domain classifier \cite{shrivastava2017cvpr,bousmalis2018roboticgrasping,li2018semantic}, as done in SimGAN \cite{shrivastava2017cvpr} and illustrated in Figure~\ref{fig:mapping1}. Alternatively, rather than learning a mapping from source to target, the opposite could be done: learn a mapping from target to source, train a classifier on the source images with known labels, and test by feeding target images to the image-to-image translation model (to make them look like source images) followed by the source-domain classifier \cite{chen2018seuda}, as illustrated in Figure~\ref{fig:mapping2}.

In either of these approaches, if the mapping and the classification models are learned independently, the class assignments may not be preserved. For instance, class 1 may end up being ``renamed'' to class 2 after the mapping since the mapping was learned ignoring the class labels. This issue can be resolved by incorporating a semantic consistency loss (see Section~\ref{losses}) and training the mapping and classification models jointly \cite{bousmalis2016nips,hoffman2018icml}, as done in PixelDA \cite{bousmalis2017cvpr}.

If there is a way to perform hyperparameter tuning, a third option is possible (combination of Figure~\ref{fig:mapping1} and \ref{fig:mapping2}): train a target-domain classifier on the source-to-target GAN (for which the GAN is not used during testing) and a source-domain classifier on the target-to-source GAN (for which the GAN is used during testing). The algorithm may then output a linear combination of the prediction results from the two classifiers \cite{russo2018cvpr}. While this approach does improve results, it requires a method of hyperparameter training (see Section~\ref{hyperparameterTuningExisting}).

All of the above approaches perform pixel-level mapping. An alternative approach is to perform feature-level mapping. Hong et al. \cite{hong2018cvpr} use a conditional GAN to learn to make the source features look more like the target features (a distinctly different idea than making the features domain invariant, which was discussed in Section~\ref{domainInvariance}). They found this particularly helpful for structured domain adaptation (e.g., semantic segmentation, in their case).

Up to this point, these domain mapping methods have used image-to-image translation to map images (or in one case features) from one domain to another and thereby improve domain adaptation performance. Another line of research using pixel-level image generation for domain adaptation is to use a GAN to generate corresponding images in multiple domains and then employ all but the last layer of the discriminator as a feature extractor for a classifier \cite{liu2016nips,mao2018unpaired}. Liu et al. \cite{liu2016nips} train a pair of GANs called CoGAN on two domains of images. Mao et al. \cite{mao2018unpaired} propose RegCGAN using only one generator and discriminator but including a domain label prepended to the input noise vector.

\subsection{Normalization Statistics} \label{normalizationStats}
Normalization layers such as batch norm \cite{ioffe2015batchnorm} are used in most neural networks \cite{santurkar2018nips}. These have benefits including allowing for higher learning rates and thus faster training \cite{ioffe2015batchnorm}, reducing initialization sensitivity \cite{ioffe2015batchnorm}, smoothing the optimization landscape and making the gradients more Lipschitz \cite{santurkar2018nips}, and allowing for deeper networks to converge \cite{wu2018groupnorm,goodfellow2016deep}. Each batch norm layer normalizes its input to have zero mean and unit variance. At test time, running averages of the batch norm parameters can be used. Alternatives have been developed including instance norm allowing use in recurrent neural networks \cite{ba2016layer} and group norm removing the dependence on batch size \cite{wu2018groupnorm}. However, none of these normalization techniques were developed with domain adaptation in mind. In the case of domain adaptation, the normalization statistics for each domain likely differ. Another line of domain adaptation research involves using per-domain batch normalization statistics.

Li et al. \cite{li2018} assume that the neural net layer weights learn task knowledge and the batch norm statistics learn domain knowledge. If this is the case, then domain adaptation can be performed by modulating all the batch norm layers' statistics from the source to target domain, a technique they call AdaBN. This has the benefit of being simple, parameter free, and complementary to other adaptation methods.

Carlucci et al. \cite{carlucci2017autodial} propose AutoDIAL, a generalization of AdaBN. In AdaBN, the target data are not used to learn the network weights but only for adjusting the batch norm statistics. AutoDIAL can utilize the target data for learning the network weights by coupling network parameters between source and target domains. They do this through adding domain alignment layers (DA-layers) that differ for source and target input data before each of the batch norm layers. Generally, batch norm computes a moving average of the statistics on a batch of the layer's input data. However, in AutoDIAL, source and target input data to DA-layers are mixed by a learnable amount before feeding this to batch norm (meaning that the batch norm statistics are now computed over some source and some target data rather than just source data or just target data). This allows the network to automatically learn how much alignment is needed at various points in the network.

\subsection{Ensemble Methods}
Given a base model such as a neural network or decision tree, an ensemble consisting of multiple models can often outperform a single model by averaging together the models' outputs (e.g., regression) or taking a vote (e.g., classification) \cite{daho2014randomforest,goodfellow2016deep}. This is because if the models are diverse then each individual model will likely make different mistakes \cite{goodfellow2016deep}. However, this performance gain corresponds with an increase in computation cost due to the large number of models to evaluate for each ensemble prediction, making ensembles common for some use cases such as competitions but uncommon when comparing models \cite{goodfellow2016deep}. Despite the incurred cost, several ensemble-based methods have been developed for domain adaptation either using the ensemble predictions to guide learning or using the ensemble to measure prediction confidence for pseudo-labeling target data.

\subsubsection{Self-Ensembling}
An alternative to using multiple instances of a base model as the ensemble is using only a single model but ``evaluating'' (via a history or average) the models in the ensemble at multiple points in time during training -- a technique called \textit{self-ensembling}. This can be done by averaging over past predictions for each example (by recording previous predictions) \cite{laine2017iclr} or past network weights (by maintaining a running average) \cite{tarvainen2017nips}. Since an ensemble requires diverse models, these self-ensembling approaches require high stochasticity in the networks, which is provided by extensive data augmentation, varying the augmentation parameters, and including dropout. These methods were originally developed for semi-supervised learning.

French et al. \cite{french2018iclr} modify and extend these prior self-ensembling methods for unsupervised domain adaptation. They use two networks: a student network and a teacher network. Input images are fed first to stochastic data augmentation (Gaussian noise, translations, horizontal flips, affine transforms, etc.) before being input to both networks. Because the method is stochastic, the augmented images fed to the networks will differ. The student network is trained with gradient descent while the teacher network weights are an exponential moving average (EMA) of the student network's weights. This method outperforms the other methods on the datasets in Table~\ref{comparePerformance1}. Athiwaratkun et al. \cite{athiwaratkun2018there} show that in at least one experiment stochastic weight averaging \cite{izmailov2018averaging} can further improve these results.

\subsubsection{Pseudo-Labeling}
Rather than voting or averaging the outputs of the models in an ensemble, the individual model predictions could be compared to determine the ensemble's confidence in that prediction. The more models in the ensemble that agree, the higher the ensemble's confidence in that prediction. In addition, if performing classification on a particular example, an individual model's confidence can be determined by looking at the last layer's softmax distribution: uniform indicates uncertainty whereas one class's probability much higher than the rest indicates higher confidence. Applying this to domain adaptation, a diverse ensemble trained on source data may be used to label target data. Then, if the ensemble is highly confident, those now-labeled target examples can be used to train a classifier for target data.

This is the method Saito et al. \cite{saito2017icml} developed called asymmetric tri-training (ATT). Two networks sharing a feature extractor are trained on the labeled source data (i.e., the ensemble in this case is of size two). Those two networks then predict the labels for the unlabeled target data, and if the two agree on the label and have high enough confidence on a particular instance, then the predicted label for that example is assumed to be the true label. After the target data are labeled by the first two networks, the third network (also sharing the same feature extractor) can be trained using the assumed-true labels (pseudo-labels). Diversity in the ensemble is handled with an additional loss (see Section~\ref{losses}).

Instead of using an ensemble, Zou et al. \cite{zou2018eccv} rely on just the softmax distribution for the confidence measure. When working with semantic segmentation, they found relying on the prediction confidence for pseudo-labeling results in transferring primarily easy classes while ignoring harder classes. Thus, they additionally propose adding a class-wise weighting term when pseudo-labeling to normalize the class-wise confidence levels and thus balance out the class distribution.

\subsection{Target Discriminative Methods} \label{clusterAssumption}
One assumption that has led to successes in semi-supervised learning algorithms is the \textit{cluster assumption} \cite{chapelle2005semi}: that data points are distributed in separate clusters and the samples in each cluster have a common label \cite{shu2018vada}. If this is the case, then decision boundaries should lie in low density regions (i.e., should not pass through regions where there are many data points) \cite{chapelle2005semi}. A variety of domain adaptation methods have been explored to move decision boundaries into density regions of lower density. These have typically been trained adversarially.

Shu et al. \cite{shu2018vada} in virtual adversarial domain adaptation (VADA) and Kumar et al. \cite{kumar2018nips} in co-regularized alignment (Co-DA) both use a combination of variational adversarial training (VAT) developed by Miyato et al. \cite{miyato2018virtual} and conditional entropy loss. They are used in combination because VAT without the entropy loss may result in overfitting to the unlabeled data points \cite{kumar2018nips} and the entropy loss without VAT may result in the network not being locally-Lipschitz and thus not resulting in moving the decision boundary away from the data points \cite{shu2018vada}. Shu et al. \cite{shu2018vada} additionally propose a decision-boundary iterative refinement step with a teacher (DIRT-T) for use after training to further refine the decision boundaries on the target data, allowing for a slight improvement over VADA. An entropy loss was also used in AutoDIAL \cite{carlucci2017autodial} but without VAT.

In generative adversarial guided learning (GAGL), Wei et al. \cite{wei2018generative} propose to let a GAN move decision boundaries into lower-density regions. Using domain alignment methods that learn domain-invariant features like DANN (Section~\ref{domainInvariance}), typically the data fed to the feature extractor is either source or target data. However, Wei et al. propose to alternate this with feeding  generated (fake) images and appending a ``fake'' label to the task classifier, thus repurposing the task classifier as a GAN discriminator. They found this to have the effect of moving the decision boundaries in the target domain into areas of lower density with a GAN, promoting target-discriminative features as a result.

Saito et al. \cite{saito2018adversarial} propose adversarial dropout regularization. Since dropout is stochastic, when they create two instances of the task classifier containing dropout, the resulting networks may produce different predictions. The difference between these predictions can be viewed as a discriminator. Using this discriminator to adversarially train the feature extractor has the effect of producing target discriminative features. Lee et al. \cite{lee2019iccv} alter adversarial dropout to better handle convolutional layers by dropping channel-wise rather than element-wise.

\subsection{Combinations} \label{combinations}
In recent work, researchers have proposed various combinations of the above methods. Domain mapping has been combined with domain-invariant feature learning methods either trained separately (in GraspGAN \cite{bousmalis2018roboticgrasping}) or jointly (in CyCADA \cite{hoffman2018icml}). Following AdaBN, many researchers started employing domain-specific batch normalization \cite{bousmalis2018roboticgrasping,french2018iclr,li2018twin,kumar2018nips,kang2019contrastive}. Kumar et al. \cite{kumar2018nips} propose co-regularized alignment (Co-DA), an approach in which two separate adversarial domain-invariant feature networks are learned with different feature spaces, drawing on ensemble-based methods. Kang et al. \cite{kang2018eccv} combine domain mapping with aligning the models' attention by minimizing an attention-based discrepancy. Deng et al. \cite{deng2019cluster} combine target discriminative methods with self-ensembling. Lee et al. \cite{lee2019sliced} combine target discriminative methods and domain-invariant feature learning with a sliced Wasserstein metric.

Multi-adversarial domain adaptation (MADA) \cite{pei2018multi} combines adversarial domain-invariant feature learning with ensemble methods for the purpose of better handling multi-modal data. This is accomplished by incorporating a separate discriminator for each class and using the task classifier's softmax probability to weight the loss from each discriminator for unlabeled target samples.

Saito et al. \cite{saito2018cvpr} combine elements of adversarial domain-invariant feature learning, ensemble methods, and target discriminative features in their maximum classifier discrepancy (MCD) method. They propose using a shared feature extractor followed by an ensemble (of size two) of task-specific classifiers, where the discrepancy between predictions measures how far outside the support of the source domain the target samples lie. The discriminator in this setup is the combination of the two classifiers. The feature extractor is trained to minimize the discrepancy (i.e., fool the classifiers that the samples are from the source domain) while the classifiers are trained to maximize the discrepancy on the target samples.

\section{Components} \label{components}
Table \ref{compare} summarizes the neural network-based domain adaptation methods we discuss showing components each method uses including what type of adaptation, which loss functions, whether the method uses a generator, and which weights are shared. Below we discuss each of these aspects followed by how the networks are trained, what types of networks can be used, multi-level adaptation techniques, and how to tune the hyperparameters of these methods.

\subsection{Losses} \label{losses}
\subsubsection{Distance}
Distance functions play a variety of roles in domain adaptation losses. A distance loss can be used to align two distributions by minimizing a distance function (e.g., MMD) as explained in Section~\ref{domainInvariance}. If using an ensemble, minimizing a distance function can align the outputs of the ensemble's models: an L1 loss of the difference in predicted target class probabilities from two networks in Co-DA \cite{kumar2018nips} or a squared difference between the predictions of the student and teacher networks in self-ensembling \cite{french2018iclr}. (Note the squared difference loss is confidence thresholded, i.e., if the max predicted output is below a certain threshold then the squared difference loss is set to zero.)

Some of the described methods have been altered replacing the task loss with one of similarity. Laradji et al. \cite{laradji2018m} propose M-ADDA, a metric-learning modification to ADDA but with the goal of maximizing the margin between clusters of data points' embeddings. Based on DANN, Pinheiro \cite{pinheiro2018cvpr} proposes SimNet, classifying based on how close an embedding is to the embeddings of a random subset of source images for each class. Hsu et al. \cite{hsu2018learning} propose $\text{CCN}^{++}$ incorporating a pairwise similarity network (trained with the same class is similar and different classes are dissimilar).

\subsubsection{Promote Differences}
Methods that rely on multiple networks learning different features (such as to make an ensemble diverse) do so by promoting differences between the networks. Saito et al. \cite{saito2017icml} train the two classifiers labeling unlabeled data to use different features by adding a norm of the product of the two classifiers' weights. Bousmalis et al. \cite{bousmalis2016nips} promote different features between two private feature extractors with a soft subspace orthogonality constraint, which is similarly used by Liu et al. \cite{liu2017adversarial} for text classification. Kumar et al. \cite{kumar2018nips} train the feature extractors to be different by pushing minibatch means apart. Saito et al. \cite{saito2018cvpr} maximize the discrepancy between two classifiers using a fixed, shared feature extractor to promote using different features.

\subsubsection{Cycle Consistency / Reconstruction}
A cycle consistency loss or reconstruction loss is commonly used in domain mapping methods to avoid requiring a dataset of corresponding images to be available in both domains. This is how CycleGAN \cite{zhu2017iccv}, DualGAN \cite{yi2017iccv}, and DiscoGAN \cite{kim2017discogan} can be unsupervised. This means that after translating an image from one domain (e.g., horses) to another (e.g., zebras), the new image can be translated back to reconstruct the original image, as illustrated in Figure~\ref{fig:CycleConsistent}. Some variants of this have been proposed such as an L1 loss with a transformation function (e.g., identity, image derivatives, mean of color channels) \cite{shrivastava2017cvpr}, a feature-level cycle-consistency loss (mapping from source to embedding to target then back to embedding resulting in the same embeddings) \cite{royer2017xgan}, or using the loss in one  \cite{choi2018cvpr} or both directions \cite{royer2017xgan,hoffman2018icml}. Sener et al. \cite{sener2016nips} enforce cycle consistency in their $k$-nearest neighbors ($k$-NN) approach by requiring the distance between any source and target point labeled the same to be less than the distance between any source and target point labeled differently and derive a rule they can solve with stochastic gradient descent.

\subsubsection{Semantic Consistency}
\begin{figure}
    \centering
    \tikzstyle{int}=[draw, minimum size=4em]
    \tikzstyle{init} = [pin edge={to-,thin,black}]

    \begin{subfigure}{0.7\textwidth}
        \centering
        \begin{tikzpicture}[node distance=3.25cm,auto,>=latex']
            \node (a) {\shortstack{\includegraphics[width=0.17\linewidth]{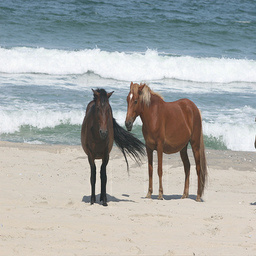} \\ \small{Source Domain} \\ \small{(Input Image)}}};
            \node (b) [right of=a] {\shortstack{\includegraphics[width=0.17\linewidth]{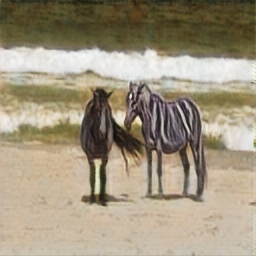} \\ \small{Mapped to} \\ \small{Target Domain}}};
            \node (c) [right of=b] {\shortstack{\includegraphics[width=0.17\linewidth]{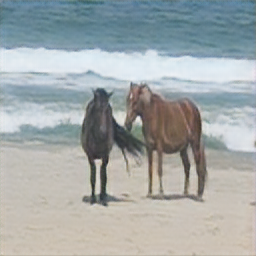} \\ \small{Mapped Back to} \\ \small{Source Domain}}};
            \node (d) [below of=b] {\shortstack{\includegraphics[width=0.17\linewidth]{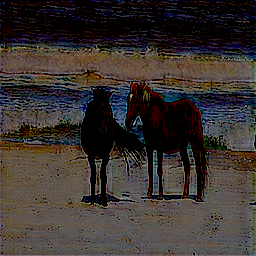} \\ \small{Difference is} \\ \small{Minimized}}};
            \path[->] (a) edge node {} (b);
            \path[->] (b) edge node {} (c);
            \path[dashed,->] (a) edge node {} (d);
            \path[dashed,->] (c) edge node {} (d);
        \end{tikzpicture}
        \caption{Cycle Consistency}
        \label{fig:CycleConsistent}
    \end{subfigure}
    \begin{subfigure}{0.29\textwidth}
        \centering
        \begin{tikzpicture}[node distance=3.5cm,auto,>=latex']
            \node (a) {\shortstack{\small{Sky} \\ \includegraphics[width=0.41\linewidth]{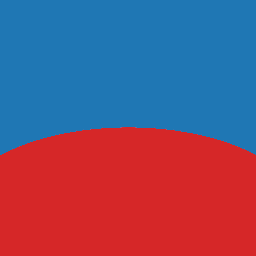} \\ \small{Ground}}};
            \node (b) [below of=a] {\shortstack{\small{Ground} \\ \includegraphics[width=0.41\linewidth]{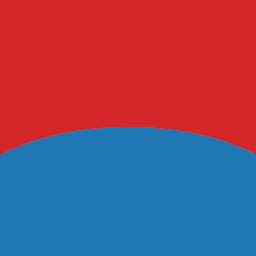} \\ \small{Sky}}};
            \path[->] (a) edge node {} (b);
        \end{tikzpicture}
        \caption{Semantic Consistency}
        \label{fig:SemanticConsistency}
    \end{subfigure}
    \caption{(a) Illustration of a cycle-consistency loss using the horses $\leftrightarrow$ zebras dataset by Zhu et al. \cite{zhu2017iccv}. The difference between the original source image and the reconstructed image (source to target and back to source) is minimized. (b) Example semantic segmentation situation in which the class names are swapped between the input image and the mapped image that would be prevented by including a semantic-consistency loss. The semantic-consistency loss requires that the class assignments are preserved.}
    \label{fig:losses}
\end{figure}
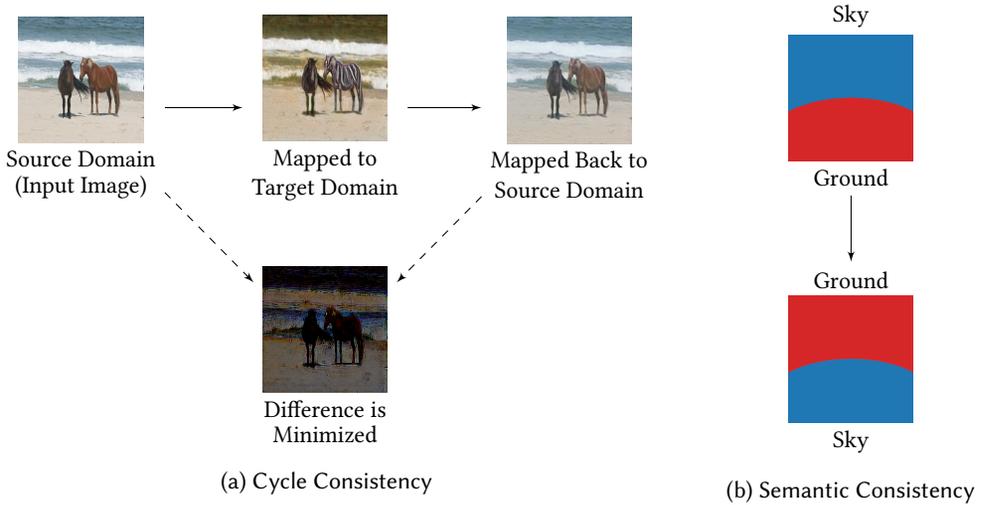

A semantic consistency loss can be used to preserve class assignments as illustrated in Figure~\ref{fig:SemanticConsistency} (a segmentation example). The semantic consistency loss requires that a classifier output (or semantic segmentation labeling) from the original source image is the same as the same classifier's output on the pixel-level mapped target output.

\subsubsection{Task}
Nearly all of the domain adaptation methods include some form of task loss that helps the network learn to perform the desired task. For example, for classification, the goal is to output the ground truth source label, or for semantic segmentation, to label each pixel with the correct ground truth source label. The task loss used is generally a cross-entropy loss, or more specifically the negative log likelihood of a softmax distribution \cite{goodfellow2016deep} when using a softmax output layer. The exceptions not including a task loss are SimNet \cite{pinheiro2018cvpr} that classify based on distance to prototypes of each class, the work by Sener et al. \cite{sener2016nips} that uses $k$ nearest neighbors, and AdaBN \cite{li2018} that only adjusts the batch norm layers to the target domain. In addition, the image-to-image translation methods are application agnostic unless trained jointly for domain adaptation.

\subsubsection{Adversarial}
A variety of methods use a discriminator (or critic) for learning domain-invariant features, realistic image generation, or promoting target discriminative features by forcing a network (either a feature extractor or generator) to produce outputs indistinguishable between two domains (source and target or real and fake). This loss is different than the other losses discussed in this section because this \textit{adversarial loss} is learned \cite{goodfellow2016survey,isola2017cvpr} (where learning is more than a hyperparameter search) rather than being provided as a predefined function. During training, gradients from the discriminator are used to train the feature extractor or generator (e.g., negated by a gradient reversal layer, Section~\ref{featureLevelAdaptation}). This alternates with updating the discriminator itself to make the correct domain classification.

\subsubsection{Additions for Specific Problems}
Some research focusing on specific problems has resulted in additional losses. For semantic segmentation, Li et al. \cite{li2018semantic} develop a loss making segmentation boundaries sharper to help when the mapped image-to-image translation images will be used for segmentation, Chen et al. \cite{chen2018cvprroad} develop a distillation loss in addition to performing location-aware alignment (e.g., ``road'' is usually at the bottom of each image), Hoffman et al. \cite{hoffman2016fcns} develop a class-aware constrained multiple instance loss, Zhang et al. \cite{zhang2017iccv} develop a curriculum where after learning some high-level properties on easy tasks the segmentation network is forced to follow those properties (interpretations include student-teacher setup or posterior regularization), and Perone et al. \cite{perone2018unsupervised} apply the self-ensembling method \cite{french2018iclr} replacing the cross-entropy loss with a consistency loss. For object detection, Chen et al. \cite{chen2018cvpr} use two domain classifiers (one on an image-level representation and the other on an instance-level representation) with a consistency regularization between them. For adaptation from synthetic images where it is known which pixels are foreground in the source images, Bousmalis et al. \cite{bousmalis2017cvpr} and Bak et al. \cite{bak2018eccv} mask certain losses to only penalize foreground pixel differences. For person re-identification, Wei et al. \cite{wei2018cvpr} include a person identity-keeping constraint in their domain mapping GAN.

\subsection{Low-Confidence or Low-Relevance Rejection} \label{rejection}
Given a measure of confidence, performance may increase if we can reject data points for training the target classifier that are not of sufficient confidence. This, of course, assumes our confidence measurement is accurate enough. Saito et al. \cite{saito2017icml} used the label agreement of an ensemble combined with the softmax distribution output (uniform is not confident, one probability much higher than the rest is confident). Sener et al. \cite{sener2016nips} used the label agreement of the $k$ nearest source data points. If the confidence is to low, then the example is rejected and not used in training until if later on when re-evaluated it is determined to be sufficiently confident. Inoue et al. \cite{inoue2018cvpr} used an object detector's prediction probability as a measure of confidence, only using high-confidence detections for fine-tuning an object detection network. Similarly, a rejection approach could be used if we have a measure of relevance. For text classification, Zhang et al. \cite{zhang2017aspect} weight examples by their relevance to their target aspect based on a small set of positive and negative keywords (a form of weak supervision).

\subsection{Weight Sharing} \label{weightSharing}
Methods employ different amounts of sharing network weights between domains or regularizing the weights to be similar. Most methods completely share weights between the feature extractors used on the source and target domains (as shown in Table~\ref{compare}). However, some techniques do not. Since deep networks consist of many layers, allowing them to represent hierarchical features, Long et al. \cite{long2015icml} propose copying the lower layers from a network trained on the source domain and adapting higher layers to the target domain with MK-MMD since higher layers do not transfer well between domains. In CoGAN, Liu et al. \cite{liu2016nips} share the first few layers of the generators and the last few layers of the discriminators, making the assumption that the domains share high-level representations. In AdaBN, Li et al. \cite{li2018} assume domain knowledge is stored in the batch norm statistics, so they share all weights except for the batch norm statistics. French et al. \cite{french2018iclr} define the teacher network as an exponential moving average of the student network's weights (a type of ensemble). Instead of sharing weights, Rozantsev et al. \cite{rozantsev2018ieee,rozantsev2018cvpr} propose two variants: regularizing weights to be similar but not penalizing linear transformations and transforming the weights from the source network to the target network with small residual networks. Bousmalis et al. \cite{bousmalis2016nips} propose domain separation networks (DSN): learning source-specific, target-specific, and shared features where the ``shared'' source domain encoder and ``shared'' target domain encoder do share weights, but the ``private'' source domain encoder and ``private'' target domain encoders do not. Others have similarly explored this idea of shared vs. specific features \cite{liu2017adversarial,ren2018factorized,cao2018dida}.

\subsection{Training Stages} \label{stages}
Some have trained networks for domain adaptation in stages. Tzeng et al. \cite{tzeng2017cvpr} train a source classifier first followed by adaptation. Taigman et al. \cite{taigman2016dtn} use a pre-trained encoder during adaptation. Bousmalis et al. \cite{bousmalis2018roboticgrasping} in GraspGAN first train the domain-mapping network followed by the domain-adversarial network. Hoffman et al. \cite{hoffman2018icml} in CyCADA train their many components in stages because it would not all fit into GPU memory at once.

Other methods train the domain adaptation networks jointly, which using an adversarial approach is done by alternating between training the discriminator and the rest of the networks (Sections~\ref{gan} and \ref{featureLevelAdaptation}). However, variations exist for some other methods. Saito et al. \cite{saito2017icml} in ATT cycle through generating training the source networks, generating pseudo-labels, and training the target network. Zou et al. \cite{zou2018eccv} alternate between pseudo-labeling the target data and re-training the model using the labels (a form of self-training). Wei et al. \cite{wei2018generative} in GAGL alternate between feeding in real source and target data and the fake images generated by a GAN. Sener et al. \cite{sener2016nips} alternate between $k$-nearest neighbors and performing gradient descent.

\subsection{Multi-Level}
Some adaptation methods perform adaptation at more than one level. As discussed in Section~\ref{combinations}, GraspGAN \cite{bousmalis2018roboticgrasping} and CyCADA \cite{hoffman2018icml} perform pixel-level adaptation with domain mapping and feature-level adaptation with domain-invariant feature learning. Hoffman et al. \cite{hoffman2018icml} found that performing both levels of adaptation significantly improves accuracy: using domain mapping to capture low-level image domain shifts and learning domain-invariant features to handle larger domain shifts than what pure domain mapping methods can support. Following this idea, Tsai et al. \cite{tsai2018cvpr} make semantic segmentation predictions and perform domain-invariant feature learning at multiple levels in their semantic segmentation network, and Zhang et al. \cite{zhang2018cvpr} perform domain-invariant feature learning at multiple levels while automatically learning how much to align to each level. Chen et al. \cite{chen2018cvpr} perform domain-invariant feature learning at both image and instance levels for object detection but also include a consistency regularization between the two domain classifiers.

\subsection{Types of Networks} \label{networkTypes}
Nearly all of the surveyed approaches focus on learning from image data and use convolutional neural networks (CNNs) such as ResNet-50 or Inception (Table~\ref{comparePerformance2}). Wang et al. \cite{wang2019transferable} explore the use of attention networks, Kang et al. \cite{kang2018eccv} a combination of CNNs and attention, Ma et al. \cite{ma2019cvpr} graph convolutional networks, and Kurmi et al. \cite{kurmi2019cvpr} Bayesian neural networks. In the case of time-series data, Purushotham et al. \cite{purushotham2017variational} propose instead using a variational recurrent neural network (RNN) \cite{NIPS2015_5653} or LSTM (a type of RNN) \cite{sepp1997lstm} rather than a CNN. The RNN learns the temporal relationships while adversarial training is used to achieve domain adaptation. For text classification (a type of natural language processing), Liu et al. \cite{liu2017adversarial} also use LSTMs while Zhang et al. \cite{zhang2017aspect} found a CNN to work just as well as RNNs or bi-LSTMs in their experiments. For relation extraction (another type of natural language processing), Fu et al. \cite{fu2017domain} also use a CNN. For time-series speech recognition, Zhao et al. \cite{zhao2017principled} use bi-LSTMs while Hosseini-Asl et al. \cite{hosseiniasl2019augmented} used a combination of CNNs and RNNs. In the related problem of domain generalization, a combination of CNNs and RNNs have been used for handling a radio spectrogram changing through time to identify sleep stages \cite{zhao2017icml}.

\begin{table*}
\caption{Comparison of different neural network based domain adaptation methods based on method of adaptation (domain-invariant feature learning [DI], domain mapping [DM], normalization [N], ensemble [En], target discriminative [TD]), various loss functions (distance, promoting different features, cycle consistency, semantic consistency, task, feature- or pixel-level adversarial), usage of a generator, and which weights are shared (in the feature extractor).}
\label{compare}

\begin{minipage}{\columnwidth}
\begin{scriptsize}
\begin{center}
{\renewcommand{\arraystretch}{1.3}
\begin{tabular}{@{}l@{}c@{}c@{}c@{}c@{\hspace{0.5em}}c@{\hspace{0.5em}}c@{\hspace{0.5em}}c@{\hspace{0.5em}}c@{}c@{}c@{}c@{}c@{}c@{}c@{}c@{}}
\toprule
\multicolumn{1}{c}{\multirow{2}{*}{\textbf{Name}}} & \multicolumn{1}{c}{\multirow{2}{*}{\textbf{Method}}} && \multicolumn{5}{c}{\textbf{Loss Functions}} && \multicolumn{2}{c}{\textbf{Adversarial Loss}} && \multicolumn{1}{c}{\multirow{2}{*}{\textbf{Generator}}} & \multicolumn{1}{c}{\multirow{2}{*}{\begin{tabular}{@{}c@{}}\textbf{Shared} \\ \textbf{Weights}\end{tabular}}} \\
\cmidrule{4-8} \cmidrule{10-11}
    & & \phantom{ab} & \textbf{Distance} & \textbf{Diff.} & \textbf{Cycle} & \textbf{Sem.} & \textbf{Task} & \phantom{ab} & \textbf{Feature} & \textbf{Pixel} & \phantom{a} && \\
\midrule

\textbf{CAN}\cite{kang2019contrastive} & DI,N && CCD & & & & \checkmark && & && & not BN \\
\hline
\textbf{French et al.}\cite{french2018iclr} & En,N && sq. diff. & & & & \checkmark && & && & EMA \\
\hline
\textbf{Co-DA}\cite{kumar2018nips}\footnote{\label{vatTraining}also incorporate virtual adversarial training \cite{miyato2018virtual}} & DI,En,N,TD && L1 & \checkmark & & & \checkmark && \checkmark & && & optional \\
\hline
\textbf{VADA}\cite{shu2018vada}\footref{vatTraining} & DI,TD && & & & & \checkmark && \checkmark & && & \checkmark \\
\hline
\textbf{DeepJDOT}\cite{damodaran2018deepjdot} & DI && JDOT & & & & \checkmark && & && & \checkmark \\
\hline
\textbf{CyCADA}\cite{hoffman2018icml} & DI,DM && & & \checkmark & \checkmark & \checkmark && \checkmark & \checkmark && \checkmark & \\
\hline
\textbf{Gen. to Adapt}\cite{sankaranarayanan2018cvpr} & DI && & & & & \checkmark && \checkmark & && \checkmark & \checkmark \\
\hline
\textbf{SimNet}\cite{pinheiro2018cvpr} & DI && prototypes & & & & && \checkmark & && & \\
\hline
\textbf{MADA}\cite{pei2018multi} & DI,En && & & & & \checkmark && \checkmark & && & \checkmark \\
\hline
\textbf{MCD}\cite{saito2018cvpr} & DI,En,TD && \checkmark & \checkmark & & & \checkmark && \checkmark & && & \\
\hline
\textbf{GAGL}\cite{wei2018generative} & DI,TD && & & & & \checkmark && \checkmark & \checkmark && \checkmark & \checkmark \\
\hline
\textbf{SBADA-GAN}\cite{russo2018cvpr}\footnote{also a self-labeled classification loss (learn label on source images, pseudo-label mapped target to source)} & DM && & & & \checkmark & \checkmark && & \checkmark && \checkmark & \\
\hline
\textbf{MCA}\cite{zhang2018mca} & DI && MCA & & & & \checkmark && & && & \checkmark \\
\hline
\textbf{$\text{CCN}^{++}$}\cite{hsu2018learning} & DI && clusters & & & & && \checkmark & && & \checkmark \\
\hline
\textbf{M-ADDA}\cite{laradji2018m} & DI && clusters & & & & \checkmark && \checkmark & && & \\
\hline
\textbf{Rozant. et al.}\cite{rozantsev2018ieee} & DI && MMD & & & & \checkmark && & && & regularize \\
\hline
\textbf{XGAN}\cite{royer2017xgan} & DM && & & \checkmark & & && \checkmark & \checkmark && \checkmark & some \\
\hline
\textbf{StarGAN}\cite{choi2018cvpr} & DM && & & \checkmark & & && & \checkmark && \checkmark & \checkmark \\
\hline
\textbf{PixelDA}\cite{bousmalis2017cvpr} & DM && & & & \checkmark & \checkmark && & \checkmark && \checkmark & \checkmark \\
\hline
\textbf{AutoDIAL}\cite{carlucci2017autodial} & N,TD && & & & & \checkmark && & && & not BN \\
\hline
\textbf{AdaBN}\cite{liu2018} & N && & & & & && & && & not BN \\
\hline
\textbf{JAN-A}\cite{long2017jmmd} & DI && JMMD & & & & \checkmark && \checkmark & && & \checkmark \\
\hline
\textbf{LogCORAL}\cite{wang2017iccv}& DI && logCOR, mean & & & & \checkmark && & && & \checkmark \\
\hline
\textbf{Log D-CORAL}\cite{morerio2017correlation}& DI && logDCOR & & & & \checkmark && & && & \checkmark \\
\hline
\textbf{VRADA}\cite{purushotham2017variational} & DI && & & & & \checkmark && \checkmark & && & \checkmark \\
\hline
\textbf{ATT}\cite{saito2017icml} & En && & \checkmark & & & \checkmark && & && & \checkmark \\
\hline
\textbf{SimGAN}\cite{shrivastava2017cvpr} & DM && & & & & && & \checkmark && \checkmark & N/A\footnote{maps to target domain so only have feature extractor for target (part of the classifier)} \\
\hline
\textbf{ADDA}\cite{tzeng2017cvpr} & DI && & & & & \checkmark && \checkmark & && & \\
\hline
\textbf{CycleGAN}\cite{zhu2017iccv} & DM && & & \checkmark & & && & \checkmark && \checkmark & \footnote{unspecified; originally not applied to domain adaptation, but later used for this \cite{hoffman2018icml,benaim2017nips,fu2018geometry}} \\
\hline
\textbf{RegCGAN}\cite{mao2018unpaired} & DM && & & & & \checkmark && \checkmark & && \checkmark & \checkmark \\
\hline
\textbf{Sener et al.}\cite{sener2016nips} & DI && $k$-NN & & & & && & && & \checkmark \\
\hline
\textbf{DSN}\cite{bousmalis2016nips} & DI && & \checkmark & \checkmark & & \checkmark && \checkmark & && & some \\
\hline
\textbf{DRCN}\cite{ghifary2016} & DI && & & \checkmark & & \checkmark && & && & \checkmark \\
\hline
\textbf{CoGAN}\cite{liu2016nips} & DM && & & & & \checkmark && \checkmark & && \checkmark & some \\
\hline
\textbf{Deep CORAL}\cite{sun2016} & DI && CORAL & & & & \checkmark && & && & \checkmark \\
\hline
\textbf{DANN}\cite{ajakan2014domain,ganin2015icml,ganin2016jmlr} & DI && & & & & \checkmark && \checkmark & && & \checkmark \\
\hline
\textbf{DAN}\cite{long2015icml} & DI && MK-MMD & & & & \checkmark && & && & low \\
\hline
\textbf{Tzeng et al.}\cite{tzeng2015iccv}\footnote{semi-supervised for some classes, i.e., requires some labeled target data for some of the classes} & DI && & & & & \checkmark && \checkmark & && & \checkmark \\

\bottomrule
\end{tabular}
}
\end{center}
\end{scriptsize}
\end{minipage}
\end{table*}

\subsection{Hyperparameter Tuning} \label{hyperparameterTuningExisting}
Normal supervised learning-based hyperparamenter tuning methods do not carry over to unsupervised domain adaptation \cite{long2013iccv,long2016nips,ganin2016jmlr,bousmalis2016nips,wang2018domain,perone2018unsupervised,morerio2018minimalentropy}. A common supervised learning approach is to split the training data into a smaller training set and a validation set. After repeatedly altering the hyperparameters, retraining the model, and testing on this validation set for each set of hyperparameters, the model yielding the highest validation set accuracy is selected. Another option is cross validation. However, in unsupervised domain adaptation, there are now two domains, and the data for the target domain may not include any labels. When evaluating domain adaptation approaches on common datasets, generally the target data does contain labels, so work by some groups \cite{bousmalis2016nips,russo2018cvpr,wang2018domain,wei2018generative,carlucci2017autodial,kumar2018nips,shu2018vada} do use some labeled target data (or all of it \cite{long2013iccv,shen2018wasserstein}) for hyperparameter tuning, which can be interpreted as an upper bound on how well the method could perform \cite{wang2018domain}. For example, some \cite{long2016nips,carlucci2017autodial} tuned for Office on one $W$ labeled example per class on the $A\rightarrow$W task, while others \cite{russo2018cvpr,wei2018generative} tuned with a validation set of 1000 randomly sampled target examples. Using any labeled target data is not ideal because real-world testing will not include labels for tuning (unless it is semi-supervised, in which case semi-supervised learning is recommended in Section~\ref{theory}).

One tuning method not requiring labeled target data is \textit{reverse validation} \cite{ganin2016jmlr}, which is a variant of \textit{reverse cross validation} \cite{zhong2010crossval}. For a set of hyperparameters, the \textit{reverse validation risk} can be estimated by first splitting source (labeled) and target (unlabeled) data into training and validation sets. Then, the labeled source and unlabeled target data are used to learn a classifier (as is normally done). Next, this forward classifier is used to label the target data and a new reverse classifier is learned (with the same algorithm) using the pseudo-labeled target data (as ``source'') and unlabeled source data (as ``target'', i.e., ignoring the known labels). This reverse classifier is evaluated on the source validation data to measure the reverse validation risk. Ganin et al. \cite{ganin2016jmlr} found this method works better if the reverse classifier is initialized with the weights of the forward classifier and if using early stopping on the source validation set and a pseudo-labeled target validation set. Finally, hyperparameters are selected (e.g., grid search, random search, Bayesian optimization, or other gradient-free optimization methods such as those implemented in Nevergrad \cite{nevergrad}) that minimize this reverse validation risk.

Alternatively, given some domain knowledge, one may devise relevant measures of similarity between the domains and tune parameters to increase the similarity. For example, French et al. \cite{french2018iclr} were able to improve performance on the challenging problem of MNIST $\rightarrow$ SVHN by tuning data augmentation hyperparameters for MNIST to match pixel intensities apparent in the SVHN dataset. By doing this, they were able to improve the state-of-the-art to 97.0\% (Table~\ref{comparePerformance1}).

\section{Results} \label{results}
Tables \ref{comparePerformance1} through \ref{comparePerformanceDatasets} summarize the results of evaluating many of these methods on datasets used for image classification as well as sentiment analysis. Care must be taken in the extent to which conclusions are drawn from comparing published numbers in different papers since the provided accuracies are for different network architectures, hyperparameters, amount of data augmentation, random initializations (or averages over a number of them), etc. and the methods may perform differently in other application areas. However, interestingly, at least one method in each of the categories of surveyed gives promising results on at least one of the datasets.

With domain-invariant feature learning with the contrastive domain discrepancy, CAN \cite{kang2019contrastive} has the highest performance on the Office dataset (Table~\ref{comparePerformance2}). By using adversarial domain-invariant feature learning, WDGRL generally outperforms the other methods on the Amazon review dataset (Table~\ref{compareSentimentPerformance}) and Generate to Adapt is second highest of the methods evaluated on the Office dataset. By using adversarial pixel-level domain mapping, SBADA-GAN \cite{russo2018cvpr} obtains the highest accuracy on MNIST$\rightarrow$MNIST-M (Table~\ref{comparePerformance1}). AutoDIAL \cite{carlucci2017autodial}, a normalization statistics method, does on-par with CAN and Generate to Adapt in two of Office adaptation tasks. The self-ensembling method by French et al. \cite{french2018iclr} outperforms all other methods on the datasets in Table~\ref{comparePerformance1}, and Co-DA \cite{kumar2018nips} comes close using an ensemble (of size two) of adversarial domain-invariant feature networks. CyCADA increases accuracy from 54\% to 82\% for a synthetic season adaptation dataset \cite{hoffman2018icml} by combining both adversarial domain-invariant feature learning and domain mapping.

A number of these promising methods use adversarial techniques, which may be a key ingredient in solving domain adaptation problems. Adversarial approaches may be helpful on certain datasets (e.g., WDGRL on the Amazon review dataset on Office), certain types of data (e.g., VRADA was developed for time series data rather than image data), or may not require as extensive of tuning (e.g., Co-DA on MNIST$\rightarrow$SVHN). Or adversarial training may be an additional tool to incorporate into existing non-adversarial methods. For instance, promising non-adversarial methods such as AutoDIAL and by French et al. could be combined with adversarial methods (see Section~\ref{combinePromisingMethods}). In fact, Long et al. \cite{long2017jmmd} develop both JAN and then the adversarial version JAN-A, and JAN-A on average outperformed JAN on the Office dataset. CAN \cite{kang2019contrastive}, which presently is the highest on the Office dataset, might also be improved by incorporating an adversarial component to it as in Long et al. \cite{long2017jmmd}.

Interestingly, French et al. by far outperform all other methods on MNIST$\rightarrow$SVHN, though this requires a problem-specific data augmentation and hyperparameter tuning. This may indicate that for some problems, maybe in particular the more challenging domain adaptation problems, hyperparameter tuning for a specific dataset may be of utmost importance. Possibly if other domain adaptation methods similarly were tuned appropriately, they would also experience large improvements. This is an area of research requiring further work (see Section~\ref{hyperparameterTuningFuture}). However, Co-DA \cite{kumar2018nips} is not far behind on SVHN$\rightarrow$MNIST and MNIST$\rightarrow$MNIST-M and is the closest on MNIST$\rightarrow$SVHN, achieving 81.7\% compared with 97.0\%. A great advantage of Co-DA is that it does not require highly-problem-specific tuning on MNIST$\rightarrow$SVHN as required by French et al. (without they only achieved 37.5\%). Possibly some components of Co-DA such as the adversarial domain adaptation or virtual adversarial training may be partially responsible for the decrease in hyperparameter sensitivity.

\begin{table*}
\caption{Classification accuracy (source $\rightarrow$ target, mean $\pm$ std \%) of different neural network based domain adaptation methods on various computer vision datasets (only including those used in $>2$ papers). Adversarial approaches denoted by $^*$.}
\label{comparePerformance1}
\begin{minipage}{\textwidth}
\begin{scriptsize}
\begin{center}
{\renewcommand{\arraystretch}{1.4}
\begin{tabular}{@{}l@{} cc @{}l@{} cc @{}l@{} c @{}l@{} cc@{}}
\toprule
\multicolumn{1}{c}{\multirow{2}{*}{\textbf{Name}}} & \multicolumn{2}{c}{\textbf{MNIST and USPS}} && \multicolumn{2}{c}{\textbf{MNIST and SVHN}} && \multicolumn{1}{c}{\textbf{MNIST[-M]}} && \multicolumn{2}{c}{\textbf{Synthetic to Real}} \\
\cmidrule{2-3} \cmidrule{5-6} \cmidrule{8-8} \cmidrule{10-11}
 & \textbf{MN $\rightarrow$ US} & \textbf{US $\rightarrow$ MN} & \phantom{a} & \textbf{SV $\rightarrow$ MN} & \textbf{MN $\rightarrow$ SV} &\phantom{a} & \textbf{MN $\rightarrow$ MN-M} &\phantom{a} & \textbf{$\text{SYN}_\text{N}$ $\rightarrow$ SV} & \textbf{$\text{SYN}_\text{S}$ $\rightarrow$ GTSRB} \\
\midrule

\multirow{3}{*}{\shortstack[l]{\textbf{Target only} \\ \textbf{(i.e., if we had} \\ \textbf{the target labels)}}} & \multirow{3}{*}{\shortstack{96.3 $\pm$ 0.1 \cite{hoffman2018icml} \\ 96.5 \cite{bousmalis2017cvpr}}} & \multirow{3}{*}{99.2 $\pm$ 0.1 \cite{hoffman2018icml}} && \multirow{3}{*}{\shortstack{99.2 $\pm$ 0.1 \cite{hoffman2018icml} \\ 99.5 \cite{bousmalis2016nips} \\ 99.51 \cite{ganin2015icml}}} &  && \multirow{3}{*}{\shortstack{96.4 \cite{bousmalis2017cvpr} \\ 98.7 \cite{bousmalis2016nips} \\ 98.91 \cite{ganin2015icml}}} && \multirow{3}{*}{\shortstack{92.44 \cite{ganin2015icml} \\ 92.4 \cite{bousmalis2016nips}}} & \multirow{3}{*}{\shortstack{99.87 \cite{ganin2015icml} \\ 99.8 \cite{bousmalis2016nips}}} \\
&&&&&&&&&&\\
&&&&&&&&&&\\
\hline
\multirow{2}{*}{\textbf{French et al.}\cite{french2018iclr}}& \multirow{2}{*}{98.2} & \multirow{2}{*}{99.5} && \multirow{2}{*}{99.3} & \multirow{2}{*}{\shortstack{37.5 \\ 97.0\footnote{problem-specific hyperparameter tuning of data augmentation to match pixel intensities of target domain images}}} &&  && \multirow{2}{*}{97.1} & \multirow{2}{*}{99.4} \\
&&&&&&&&&&\\
\hline
\textbf{Co-DA}\cite{kumar2018nips}\footnote{\label{hyperparamTunedTarget}hyperparameter tuned on some labeled target data}$^*$ & & && 98.6 & 81.7 && 97.5 && 96.0 & \\
\hline
\textbf{DIRT-T}\cite{shu2018vada}\footref{hyperparamTunedTarget}$^*$ & & && 99.4 & 76.5 && 98.7 && 96.2 & 99.6 \\
\hline
\textbf{VADA}\cite{shu2018vada}\footref{hyperparamTunedTarget}$^*$ & & && 94.5 & 73.3 && 95.7 && 94.9 & 99.2 \\
\hline
\textbf{DeepJDOT}\cite{damodaran2018deepjdot} & 95.7 & 96.4 && 96.7 & && 92.4 && & \\
\hline
\textbf{CyCADA}\cite{hoffman2018icml}$^*$ & 95.6 $\pm$ 0.2 & 96.5 $\pm$ 0.1 && 90.4 $\pm$ 0.4 &  &&  &&  &  \\
\hline
\textbf{Gen. to Adapt}\cite{sankaranarayanan2018cvpr}$^*$ & 92.8 $\pm$ 0.9 & 90.8 $\pm$ 1.3 && 92.4 $\pm$ 0.9 &  &&  &&  &  \\
\hline
\textbf{SimNet}\cite{pinheiro2018cvpr}$^*$ & 96.4 & 95.6 &&  &  && 90.5 &&  &  \\
\hline
\textbf{MCD}\cite{saito2018cvpr}$^*$ & 96.5 $\pm$ 0.3 & 94.1 $\pm$ 0.3 && 96.2 $\pm$ 0.4 & && && & 94.4 $\pm$ 0.3 \\
\hline
\textbf{GAGL}\cite{wei2018generative}\footref{hyperparamTunedTarget}$^*$ &  &  && 96.7 & 74.6 && 94.9 && 93.1 & 97.6 \\
\hline
\textbf{SBADA-GAN}\cite{russo2018cvpr}\footref{hyperparamTunedTarget}$^*$ & 97.6 & 95.0 && 76.1 & 61.1 && 99.4 &&  & 96.7 \\
\hline
\textbf{MCA}\cite{zhang2018mca} & & && 96.6 &  && 96.8 && 89.0 &  \\
\hline
\textbf{$\text{CCN}^{++}$}\cite{hsu2018learning}$^*$ & & && 89.1 &  &&  &&  &  \\
\hline
\textbf{M-ADDA}\cite{laradji2018m}$^*$ & 98 & 97 && &  &&  &&  &  \\
\hline
\textbf{Rozantsev et al.}\cite{rozantsev2018ieee} & 60.7 & 67.3 &&  &  &&  &&  &  \\
\hline
\textbf{PixelDA}\cite{bousmalis2017cvpr}$^*$ & 95.9 &  &&  &  && 98.2 &&  &  \\
\hline
\textbf{ATT}\cite{saito2017icml} &  &  && 85.0 & 52.8 && 94.0 && 92.9 & 96.2 \\
\hline
\textbf{ADDA}\cite{tzeng2017cvpr}$^*$ & 89.4 $\pm$ 0.2 & 90.1 $\pm$ 0.8 && 76.0 $\pm$ 1.8 &  &&  &&  &  \\
\hline
\textbf{RegCGAN}\cite{mao2018unpaired}$^*$ & 93.1 $\pm$ 0.7 & 89.5 $\pm$ 0.9 && & && && &  \\
\hline
\textbf{DTN}\cite{taigman2016dtn}$^*$ & & && 84.4 & && && &  \\
\hline
\textbf{Sener et al.}\cite{sener2016nips} & & && 78.8 & 40.3 && 86.7 && &  \\
\hline
\textbf{DSN}\cite{bousmalis2016nips}\footref{hyperparamTunedTarget}$^*$ & 91.3 \cite{bousmalis2017cvpr} &  && 82.7 &  && 83.2 && 91.2 & 93.1 \\
\hline
\textbf{DRCN}\cite{ghifary2016} & 91.80 $\pm$ 0.09 & 73.67 $\pm$ 0.04 && 81.97 $\pm$ 0.16 & 40.05 $\pm$ 0.07 &&  &&  &  \\
\hline
\textbf{CoGAN}\cite{liu2016nips}$^*$ & 91.2 $\pm$ 0.8 & 89.1 $\pm$ 0.8 &&  &  && 62.0 \cite{bousmalis2017cvpr} &&  &  \\
\hline
\multirow{4}{*}{\textbf{DANN}\cite{ganin2015icml,ganin2016jmlr}$^*$} & \multirow{4}{*}{85.1 \cite{bousmalis2017cvpr}} &  && \multirow{4}{*}{\shortstack{71.07 \\ 70.7 \cite{bousmalis2016nips} \\ 71.1 \cite{saito2017icml} \\ 73.6 \cite{hoffman2018icml}}} & \multirow{4}{*}{35.7 \cite{saito2017icml}} && \multirow{4}{*}{\shortstack{81.49 \\ 77.4 \cite{bousmalis2016nips} \\ 81.5 \cite{saito2017icml}}} && \multirow{4}{*}{\shortstack{90.48 \\ 90.3 \cite{bousmalis2016nips,saito2017icml}}} & \multirow{4}{*}{\shortstack{88.66 \\ 88.7 \cite{saito2017icml} \\ 92.9 \cite{bousmalis2016nips}}} \\
&&&&&&&&&&\\
&&&&&&&&&&\\
&&&&&&&&&&\\
\hline
\multirow{4}{*}{\textbf{DAN}\cite{long2015icml}} & \multirow{4}{*}{81.1 \cite{bousmalis2017cvpr}} &  && \multirow{4}{*}{\shortstack{71.1 \cite{bousmalis2016nips}}} &  && \multirow{4}{*}{\shortstack{76.9 \cite{bousmalis2016nips}}} && \multirow{4}{*}{\shortstack{88.0 \cite{bousmalis2016nips}}} & \multirow{4}{*}{\shortstack{91.1 \cite{bousmalis2016nips}}} \\
&&&&&&&&&&\\
&&&&&&&&&&\\
&&&&&&&&&&\\
\hline
\multirow{3}{*}{\shortstack[l]{\textbf{Source only} \\ \textbf{(i.e., no adaptation)}}} & \multirow{3}{*}{\shortstack{78.9 \cite{bousmalis2017cvpr} \\ 82.2 $\pm$ 0.8 \cite{hoffman2018icml}}} & \multirow{3}{*}{69.6 $\pm$ 3.8 \cite{hoffman2018icml}} && \multirow{3}{*}{\shortstack{59.19 \cite{ganin2015icml} \\ 59.2 \cite{bousmalis2016nips} \\ 67.1 $\pm$ 0.6 \cite{hoffman2018icml} }} &  && \multirow{3}{*}{\shortstack{56.6 \cite{bousmalis2016nips} \\ 57.49 \cite{ganin2015icml} \\ 63.6 \cite{bousmalis2017cvpr}}} && \multirow{3}{*}{\shortstack{86.65 \cite{ganin2015icml} \\ 86.7 \cite{bousmalis2016nips}}} & \multirow{3}{*}{\shortstack{74.00 \cite{ganin2015icml} \\ 85.1 \cite{bousmalis2016nips}}} \\
&&&&&&&&&&\\
&&&&&&&&&&\\

\bottomrule
\end{tabular}
}
\end{center}
\end{scriptsize}
\end{minipage}
\end{table*}

\begin{table*}
\caption{Classification accuracy (source $\rightarrow$ target, mean $\pm$ std \%) of different neural network based domain adaptation methods on the Office computer vision dataset. Adversarial approaches denoted by $^*$.}
\label{comparePerformance2}
\begin{minipage}{\textwidth}
\begin{scriptsize}
\begin{center}
{\renewcommand{\arraystretch}{1.4}
\begin{tabular}{@{}l cccccc@{}}
\toprule
\multicolumn{1}{c}{\multirow{2}{*}{\textbf{Name}}} & \multicolumn{6}{c}{\textbf{Office (Amazon, DSLR, Webcam)}} \\
\cmidrule{2-7}
 & \textbf{A $\rightarrow$ W} & \textbf{D $\rightarrow$ W} & \textbf{W $\rightarrow$ D} & \textbf{A $\rightarrow$ D} & \textbf{D $\rightarrow$ A} & \textbf{W $\rightarrow$ A} \\
\midrule

\textbf{CAN}\cite{kang2019contrastive}\footnote{\label{renset50Net}with ResNet-50 network} & 94.5 $\pm$ 0.3 & 99.1 $\pm$ 0.2 & 99.8 $\pm$ 0.2 & 95.0 $\pm$ 0.3 & 78.0 $\pm$ 0.3 & 77.0 $\pm$ 0.3 \\
\hline
\textbf{Gen. to Adapt}\cite{sankaranarayanan2018cvpr}\footref{renset50Net}$^*$ & 89.5 $\pm$ 0.5 & 97.9 $\pm$ 0.3 & 99.8 $\pm$ 0.4 & 87.7 $\pm$ 0.5 & 72.8 $\pm$ 0.3 & 71.4 $\pm$ 0.4 \\
\hline
\textbf{SimNet}\cite{pinheiro2018cvpr}\footref{renset50Net}$^*$ & 88.6 $\pm$ 0.5 & 98.2 $\pm$ 0.2 & 99.7 $\pm$ 0.2 & 85.3 $\pm$ 0.3 & 73.4 $\pm$ 0.8 & 71.8 $\pm$ 0.6 \\
\hline
\textbf{MADA}\cite{pei2018multi}\footref{renset50Net}$^*$ & 90.0 $\pm$ 0.1 & 97.4 $\pm$ 0.1 & 99.6 $\pm$ 0.1 & 87.8 $\pm$ 0.2 & 70.3 $\pm$ 0.3 & 66.4 $\pm$ 0.3 \\
\hline
\textbf{AutoDIAL}\cite{carlucci2017autodial}\footnote{\label{inceptionNet}with Inception-based network}\footnote{hyperparameter tuned on one $W$ labeled example per class on $A\rightarrow$W task (see \cite{long2016nips})} & 84.2 & 97.9 & 99.9 & 82.3 & 64.6 & 64.2 \\
\hline
\textbf{$\text{CCN}^{++}$}\cite{hsu2018learning}\footnote{\label{resnet18Net}with ResNet-18 network}$^*$ & 78.2 & 97.4 & 98.6 & 73.5 & 62.8 & 60.6 \\
\hline
\textbf{Rozantsev et al.}\cite{rozantsev2018ieee} & 76.0 & 96.7 & 99.6 &  &  &  \\
\hline
\textbf{AdaBN}\cite{liu2018}\footref{inceptionNet} & 74.2 & 95.7 & 99.8 & 73.1 & 59.8 & 57.4 \\
\hline
\textbf{JAN-A}\cite{long2017jmmd}\footref{renset50Net}$^*$ & 86.0 $\pm$ 0.4 & 96.7 $\pm$ 0.3 & 99.7 $\pm$ 0.1 & 85.1 $\pm$ 0.4 & 69.2 $\pm$ 0.4 & 70.7 $\pm$ 0.5 \\
\hline
\textbf{LogCORAL}\cite{wang2017iccv} & 70.2 $\pm$ 0.6 & 95.5 $\pm$ 0.1 & 99.5 $\pm$ 0.3 & 69.4 $\pm$ 0.5 & 51.2 $\pm$ 0.3 & 51.6 $\pm$ 0.5 \\
\hline
\textbf{Log D-CORAL}\cite{morerio2017correlation} & 68.5 & 95.3 & 98.7 & 62.0 & 40.6 & 40.6 \\
\hline
\textbf{ADDA}\cite{tzeng2017cvpr}\footref{renset50Net}$^*$ & 75.1 & 97.0 & 99.6 &  &  &  \\
\hline
\textbf{Sener et al.}\cite{sener2016nips} & 81.1 & 96.4 & 99.2 & 84.1 & 58.3 & 63.8 \\
\hline
\textbf{DRCN}\cite{ghifary2016} & 68.7 $\pm$ 0.3 & 96.4 $\pm$ 0.3 & 99.0 $\pm$ 0.2 & 66.8 $\pm$ 0.5 & 56.0 $\pm$ 0.5 & 54.9 $\pm$ 0.5 \\
\hline
\textbf{Deep CORAL}\cite{sun2016} & 66.4 $\pm$ 0.4 & 95.7 $\pm$ 0.3 & 99.2 $\pm$ 0.1 & 66.8 $\pm$ 0.6 & 52.8 $\pm$ 0.2 & 51.5 $\pm$ 0.3 \\
\hline
\multirow{4}{*}{\textbf{DANN}\cite{ganin2015icml,ganin2016jmlr}$^*$} & \multirow{4}{*}{\shortstack{67.3 $\pm$ 1.7 \\ 72.6 $\pm$ 0.3 \cite{ghifary2016} \\ 73.0 \cite{rozantsev2018ieee,tzeng2017cvpr}}} & \multirow{4}{*}{\shortstack{94.0 $\pm$ 0.8 \\ 96.4 $\pm$ 0.1 \cite{ghifary2016} \\ 96.4 \cite{rozantsev2018ieee,tzeng2017cvpr}}} & \multirow{4}{*}{\shortstack{93.7 $\pm$ 1.0 \\ 99.2 $\pm$ 0.3 \cite{ghifary2016} \\ 99.2 \cite{rozantsev2018ieee,tzeng2017cvpr}}} & \multirow{4}{*}{67.1 $\pm$ 0.3 \cite{ghifary2016}} & \multirow{4}{*}{54.5 $\pm$ 0.4 \cite{ghifary2016}} & \multirow{4}{*}{52.7 $\pm$ 0.2 \cite{ghifary2016}} \\
&&&&&&\\
&&&&&&\\
&&&&&&\\
\hline
\multirow{4}{*}{\textbf{DAN}\cite{long2015icml}} & \multirow{4}{*}{\shortstack{68.5 $\pm$ 0.4 \\ 63.8 $\pm$ 0.4 \cite{sun2016} \\ 64.5 \cite{rozantsev2018ieee} \\ 68.5 \cite{tzeng2017cvpr}}} & \multirow{4}{*}{\shortstack{96.0 $\pm$ 0.3 \\ 94.6 $\pm$ 0.5 \cite{sun2016} \\ 95.2 \cite{rozantsev2018ieee} \\ 96.0 \cite{tzeng2017cvpr}}} & \multirow{4}{*}{\shortstack{99.0 $\pm$ 0.2 \\ 98.6 \cite{rozantsev2018ieee} \\ 98.8 $\pm$ 0.6 \cite{sun2016} \\ 99.0 \cite{tzeng2017cvpr}}} & \multirow{4}{*}{\shortstack{67.0 $\pm$ 0.4 \\ 65.8 $\pm$ 0.4 \cite{sun2016}}} & \multirow{4}{*}{\shortstack{54.0 $\pm$ 0.4 \\ 52.8 $\pm$ 0.4 \cite{sun2016}}} & \multirow{4}{*}{\shortstack{53.1 $\pm$ 0.3 \\ 51.9 $\pm$ 0.5 \cite{sun2016}}} \\
&&&&&&\\
&&&&&&\\
&&&&&&\\
\hline
\textbf{Tzeng et al.}\cite{tzeng2015iccv} \footnote{semi-supervised for some classes, but evaluated on 16 hold-out categories for which the labels were not seen during training}$^*$ & 59.3 $\pm$ 0.6 & 90.0 $\pm$ 0.2 & 97.5 $\pm$ 0.1 & 68.0 $\pm$ 0.5 & 43.1 $\pm$ 0.2 & 40.5 $\pm$ 0.2 \\
\hline
\multirow{3}{*}{\shortstack[l]{\textbf{Source only} \\ \textbf{(i.e., no adaptation)}}} & \multirow{3}{*}{62.6 \cite{tzeng2017cvpr}\footref{renset50Net}} & \multirow{3}{*}{96.1 \cite{tzeng2017cvpr}\footref{renset50Net}} & \multirow{3}{*}{98.6 \cite{tzeng2017cvpr}\footref{renset50Net}} &  &  &  \\
&&&&&&\\
&&&&&&\\

\bottomrule
\end{tabular}
}
\end{center}
\end{scriptsize}
\end{minipage}
\end{table*}

\begin{table*}
\begin{minipage}{\columnwidth}
\caption[Sentiment Analysis]{Classification accuracy comparison for domain adaptation methods for sentiment analysis (positive or negative review) on the Amazon review dataset \cite{blitzer2007biographies}\footnote{\url{http://www.cs.jhu.edu/~mdredze/datasets/sentiment/}} with domains books (B), DVD (D), electronics (E), and kitchen (K). Adversarial approaches denoted by $^*$.}
\label{compareSentimentPerformance}
\begin{scriptsize}
\begin{center}
{\renewcommand{\arraystretch}{1.3}
\begin{tabular}{@{}c@{}c@{}cccccc@{}}
\toprule
\textbf{Source $\rightarrow$ Target} &\phantom{ab}& \textbf{DANN}\cite{ganin2016jmlr}\footnote{using 30,000-dimensional feature vectors from marginalized stacked denoising autoencoders (mSDA) by Chen et al. \cite{chen2012marginalized}, which is an unsupervised method of learning a feature representation from the training data}$^*$ & \textbf{DANN}\cite{ganin2016jmlr}\footnote{\label{features5000}using 5000-dimensional unigram and bigram feature vectors}$^*$ & \textbf{CORAL}\cite{sun2016aaai}\footnote{using bag-of-words feature vectors including only the top 400 words, but suggest using deep text features in future work} & \textbf{ATT}\cite{saito2017icml}\footref{features5000} & \textbf{WDGRL}\cite{shen2018wasserstein}\footref{features5000}\footnote{the best results on target data for various hyperparameters}$^*$ & \textbf{No Adapt.}\cite{sun2016aaai}\footnote{using bag-of-words feature vectors} \\
\midrule

\textbf{B $\rightarrow$ D} && 82.9 & 78.4 &  & 80.7 & 83.1 & \\
\hline
\textbf{B $\rightarrow$ E} && 80.4 & 73.3 & 76.3 & 79.8 & 83.3 & 74.7 \\
\hline
\textbf{B $\rightarrow$ K} && 84.3 & 77.9 &  & 82.5 & 85.5 &  \\
\hline
\textbf{D $\rightarrow$ B} && 82.5 & 72.3 & 78.3 & 73.2 & 80.7 & 76.9 \\
\hline
\textbf{D $\rightarrow$ E} && 80.9 & 75.4 &  & 77.0 & 83.6 &  \\
\hline
\textbf{D $\rightarrow$ K} && 84.9 & 78.3 &  & 82.5 & 86.2 &  \\
\hline
\textbf{E $\rightarrow$ B} && 77.4 & 71.3 &  & 73.2 & 77.2 &  \\
\hline
\textbf{E $\rightarrow$ D} && 78.1 & 73.8 &  & 72.9 & 78.3 &  \\
\hline
\textbf{E $\rightarrow$ K} && 88.1 & 85.4 & 83.6 & 86.9 & 88.2 & 82.8 \\
\hline
\textbf{K $\rightarrow$ B} && 71.8 & 70.9 &  & 72.5 & 77.2 &  \\
\hline
\textbf{K $\rightarrow$ D} && 78.9 & 74.0 & 73.9 & 74.9 & 79.9 & 72.2 \\
\hline
\textbf{K $\rightarrow$ E} && 85.6 & 84.3 &  & 84.6 & 86.3 &  \\
\bottomrule

\end{tabular}
}
\end{center}
\end{scriptsize}
\end{minipage}
\end{table*}

\begin{table*}
\caption{List and description of computer vision datasets from Tables \ref{comparePerformance1} and \ref{comparePerformance2}}
\label{comparePerformanceDatasets}
\begin{minipage}{\columnwidth}
\begin{scriptsize}
\begin{center}
{\renewcommand{\arraystretch}{1.3}
\begin{tabular}{@{}ll@{}}
\toprule
\multicolumn{2}{c}{\textbf{Computer Vision Datasets used for Domain Adaptation}} \\
\midrule

\textbf{MNIST}\cite{lecun1998mnist}\footnote{\url{http://yann.lecun.com/exdb/mnist/}} & \multicolumn{1}{p{0.8\columnwidth}}{This is a binary (mostly black and white, but actually grayscale due to anti-aliasing) handwritten digit dataset (digits 0-9), which stands for ``modified NIST.'' It is based on the National Institute of Standards and Technology's (NIST) Special Database 1 and 3, one of which was easier than the other, so MNIST is a combination of the two that are size normalized to fit in a 20x20 box preserving the aspect ratio and centered in a 28x28 pixel image.} \\

\textbf{MNIST-M}\cite{ganin2016jmlr}\footnote{\label{ganinsite}See Ganin's website \url{http://yaroslav.ganin.net/} for links to download.} & \multicolumn{1}{p{0.8\columnwidth}}{This is a modification of MNIST where the digits are blended with random patches from BSDS500 dataset color photos.} \\

\textbf{USPS}\cite{le1990handwritten}\footnote{This can be found on various sites and some Github repositories. One such place: \url{https://web.stanford.edu/~hastie/ElemStatLearn/data.html}} & \multicolumn{1}{p{0.8\columnwidth}}{This is another handwritten digit dataset (digits 0-9). It consists of handwritten zipcodes scanned and segmented by the U.S. Postal Service (USPS). They were size normalized to 16x16 pixels preserving the aspect ratio. The values are normalized to be between -1 and 1.} \\

\textbf{SVHN}\cite{netzer2011reading}\footnote{\url{http://ufldl.stanford.edu/housenumbers}} & \multicolumn{1}{p{0.8\linewidth}}{The Streetview House Numbers (SVHN) consists of single digits extracted from images of urban house numbers in Google Street View. The digits have been size normalized to 32x32 pixels.} \\

\textbf{$\text{SYN}_\text{N}$}\cite{ganin2016jmlr}\footref{ganinsite} & \multicolumn{1}{p{0.8\columnwidth}}{Ganin et al. \cite{ganin2016jmlr} used Microsoft Windows fonts to create a synthetic digit dataset (``Syn Numbers'') consisting of 1-3 digit numbers with various positions, orientation, background color, stroke color, and amount of blur. } \\

\textbf{$\text{SYN}_\text{S}$}\cite{moiseev2013evaluation}\footnote{The synthetic dataset linked to on: \url{http://graphics.cs.msu.ru/en/research/projects/imagerecognition/trafficsign}} & \multicolumn{1}{p{0.8\columnwidth}}{This is a synthetic sign dataset created from modifications to Wikipedia pictograms of traffic signs. It consists of 100,000 images and 43 classes of signs.} \\

\textbf{GTSRB}\cite{Stallkamp-IJCNN-2011}\footnote{\url{http://benchmark.ini.rub.de/?section=gtsrb&subsection=dataset}} & \multicolumn{1}{p{0.8\columnwidth}}{The German Traffic Signs Recognition Benchmark (GTSRB) is a dataset created from video taken driving around Germany. It consists of about 50,000 images and 43 classes of signs.} \\

\textbf{Office}\cite{saenko2010adapting}\footnote{\url{http://ai.bu.edu/adaptation.html}} & \multicolumn{1}{p{0.8\columnwidth}}{This dataset consists of 31 classes of objects in three different domains: Amazon (taken from its online website; medium resolution and studio lighting), DSLR (taken with a digital SLR camera; high resolution and in a real-world environment), and Webcam (taken with a 640x480 computer webcam; have noise, artifacts, and white balance issues). Note: due to Office's small size, some networks \cite{ganin2016jmlr,sun2016,rozantsev2018ieee} were pre-trained on ImageNet.} \\
\bottomrule

\end{tabular}
}
\end{center}
\end{scriptsize}
\end{minipage}
\end{table*}

\section{Theory} \label{theory}
Having surveyed domain adaptation methods, we now address the question of when adaptation may be beneficial. Ben-David et al. \cite{ben2010ml} develop a theory answering this in terms of an ideal predictor on both domains, Zhao et al. \cite{zhao2019learning} further this theory by removing the dependence on a joint ideal predictor while focusing on domain-invariant feature learning methods, and Le et al. \cite{le2018theoretical} develop theory looking beyond domain-invariant methods. These theoretical results can help answer two questions: (1) when will a classifier (or other predictor) trained on the source data perform well on the target data, and (2) given a small number of labeled target examples, how can they best be used during training to minimize target test error?

Answering the first question, labeled source data and unlabeled target data are both required (unsupervised). Answering the second question, additionally some labeled target data are required (semi-supervised). We will first review the theoretical bounds followed by a discussion of what insights these bounds provide into answering the above two questions. Ben-David et al. \cite{ben2010ml} also address the case of multiple source domains, as do Mansour et al. \cite{mansour2009nips}. In this paper, we have focused on the cases containing only one source and one target (as is common in the methods we survey).

\subsection{Unsupervised}
\subsubsection{Shared Hypothesis Space}
Ben-David et al. \cite{ben2010ml} propose setting a bound on the target error based on the source error and the divergence between the source and target domains. The empirical source error is easy to obtain by first training and then testing a classifier. However, the divergence between the domains cannot be directly obtained with standard methods like Kullback-Leibler divergence due to only having a finite number of samples from the domains and not assuming any particular distribution. Thus, an alternative is to measure it using a classifier-induced divergence called $\mathcal{H} \Delta \mathcal{H}$-divergence. Estimates of this divergence with finite samples converges to the real $\mathcal{H} \Delta \mathcal{H}$-divergence. This divergence can be estimated by measuring the error when getting a classifier to discriminate between the unlabeled source and target examples; though, it is often intractable to find the theoretically-required divergence upper bound. Using the empirical source error $\hat{\epsilon}_S(h)$, the $\mathcal{H} \Delta \mathcal{H}$-divergence between source and target samples $d_{\mathcal{H} \Delta \mathcal{H}}(\mathcal{\hat{D}}_S, \mathcal{\hat{D}}_T)$, and ideal predictor error $\lambda^*$ using the optimal hypothesis for the source and target, the target error $\epsilon_T(h)$ can be bounded as shown in Equation~\ref{theory1} (using the form given by Zhao et al. \cite{zhao2019learning}), $\forall h \in \mathcal{H}$ with probability at least $1-\delta$ for $\delta \in (0,1)$.
\begin{equation}
\epsilon_T(h) \leq \hat{\epsilon}_S(h) + \frac{1}{2} d_{\mathcal{H} \Delta \mathcal{H}}(\mathcal{\hat{D}}_S, \mathcal{\hat{D}}_T) + \lambda^* + O \left( \sqrt{\frac{d \log n + \log(\frac{1}{\delta})}{n}} \right)
\label{theory1}
\end{equation}

Zhao et al. \cite{zhao2019learning} develop another upper bound that removes the reliance on $\lambda^*$. Let $\mathcal{H} \subseteq [0,1]^\mathcal{X}$, $\mathcal{\tilde{H}} \coloneqq \{\text{sgn} \left( |h(x) - h'(x)| - t \right) | h,h' \in \mathcal{H}, 0 \leq t \leq 1 \}$, $\langle \mathcal{D}_S, f_S \rangle$ and $\langle \mathcal{D}_T, f_T \rangle$ be the source and target domains (the true distributions, not empirical). The target error can then be bounded by the source error $\epsilon_S(h)$, the discrepancy between marginal distributions $d_\mathcal{\tilde{H}}(\mathcal{D}_S, \mathcal{D}_T)$, and the distance between the optimal source and target labeling functions $\forall h \in \mathcal{H}$, as shown in Equation~\ref{theory2}.
\begin{equation}
    \epsilon_T(h) \leq \epsilon_S(h) + d_\mathcal{\tilde{H}}(\mathcal{D}_S, \mathcal{D}_T) + \min\{ \mathbb{E}_{\mathcal{D}_S}[|f_S-f_T|], \mathbb{E}_{\mathcal{D}_T}[|f_S-f_T|] \}
\label{theory2}
\end{equation}

Zhao et al. \cite{zhao2019learning} also develop an information-theoretic lower bound for target error. Let the labeling function $Y = f(X) \in \{0,1\}$, the prediction function $\hat{Y} = h(g(X)) \in \{0,1\}$, and $Z$ be the intermediate representation output by a shared feature extractor used on source and target domain data. If the Jensen-Shannon distance $d_{JS}(\mathcal{D}_S^Y, \mathcal{D}_T^Y) \geq d_{JS}(\mathcal{D}_S^Z, \mathcal{D}_T^Z)$ and the Markov chain $X \xrightarrow{g} Z \xrightarrow{h} \hat{Y}$ holds, then Equation~\ref{theory3} provides a lower bound on the source and target error.
\begin{equation}
    \epsilon_S(h \circ g) + \epsilon_T(h \circ g) \geq \frac{1}{2}\left( d_{JS}(\mathcal{D}_S^Y, \mathcal{D}_T^Y) - d_{JS}(\mathcal{D}_S^Z, \mathcal{D}_T^Z) \right)^2
\label{theory3}
\end{equation}

\subsubsection{Different Hypothesis Spaces}
Le et al. \cite{le2018theoretical} develop an upper bound that allows for different hypothesis spaces for source and target functions, possibly non-deterministic labeling, and any bounded or continuous loss. If $l$ is a bounded or continuous loss, $x \sim \mathbb{P}^s$ (source) and $x \sim \mathbb{P}^t$ (target), $T : \mathcal{X}^s \rightarrow \mathcal{X}^t$ and $K \coloneqq T^{-1}$ (bijective mapping), $R(\theta) = \mathbb{E}_{p(x,y)} [ l(y,h_\theta(x)) ]$ for $\theta$ parameterizing a hypothesis set $\mathcal{H} = \{ h_\theta | \theta \in \Theta \}$, $\Delta R(h^s, h^t) \coloneqq |R^t(h^t) - R^s(h^s)|$, $y \in \{-1,1\}$, $M$ is the number of labels, $\mathbb{P}^\# \coloneqq K_\# \mathbb{P}^t$ is the pushforward probability distribution transporting $\mathbb{P}^t$ via $K$, $\Delta p(y|x) \coloneqq p^t(y|T(x)) - p^s(y|x)$ for the true source and target labeling functions $p^s(y|x)$ and $p^t(y|x)$, where ${WS}_c(\mathbb{P}^s,\mathbb{P}^\#)$ denotes the Wasserstein-1 distance between the source and target distributions with a cost function $c(x,x') = 1_{x \neq x'}$ (1 if $x \neq x'$, otherwise 0), then Equation~\ref{theory4} provides an upper bound for the variance between a general loss on the source and target predictions.
\begin{equation}
    \Delta R(h^s, h^t) \leq M \left( {WS}_c(\mathbb{P}^s,\mathbb{P}^\#) + \min\{ \mathbb{E}_{\mathbb{P}^\#} [ \| \Delta p(y|x) \|_1 ], \mathbb{E}_{\mathbb{P}^s} [ \| \Delta p(y|x) \|_1 ]\} \right)
\label{theory4}
\end{equation}

\subsection{Semi-Supervised}
In the semi-supervised case, a linear combination of the source and target errors is computed \cite{ben2010ml}, called the $\alpha$-error. A bound can be calculated on the true $\alpha$-error based on the empirical $\alpha$-error. Finding the minimum $\alpha$-error depends on the empirical $\alpha$-error, the divergence between source and target, and the number of labeled source and target examples. Experimentation can be used to empirically determine the values of $\alpha$ that will perform well. Ben-David et al. \cite{ben2010ml} also demonstrate the process on sentiment classification, illustrating that the optimum uses non-trivial values.

The bound is given in Equation \ref{benLabelBound}. If $S$ is a labeled sample of size $m$ with $(1-\beta) m$ points drawn from the source distribution and $\beta m$ from the target distribution, then with at least probability $1-\delta$ for $\delta \in (0,1)$:
\begin{align}
\epsilon_T(\hat{h}) \leq
	& \epsilon_T(h^*_T) + 4 \sqrt{\frac{\alpha^2}{\beta} + \frac{(1-\alpha)^2}{1-\beta}} \sqrt{\frac{2d \log(2(m+1)) + 2 \log(\frac{8}{\delta})}{m}} + \nonumber \\
	& 2(1-\alpha) \left( \frac{1}{2} \hat{d}_{\mathcal{H} \Delta \mathcal{H}}(\mathcal{U}_S, \mathcal{U}_T) + 4 \sqrt{\frac{2d \log(2m') + \log(\frac{8}{\delta})}{m'}} + \lambda \right)
\label{benLabelBound}
\end{align}

Here, $\hat{h} \in \mathcal{H}$ is the empirical minimizer of the $\alpha$-error on $S$ given by $\hat{\epsilon}_\alpha(h) = \alpha \hat{\epsilon}_T(h) + (1-\alpha) \hat{\epsilon}_S(h)$ and $h^*_T = \min_{h \in \mathcal{H}} \epsilon_T(h)$ is the target error minimizer.

The optimum $\alpha$ is then:
\begin{align}
\alpha^*(m_T,m_S;D) =
\begin{cases}
	1 & m_T \geq D^2 \\
	\min\{ 1,\nu \} & m_T \leq D^2
\end{cases}
\end{align}

Here, $m_S = (1-\beta)m$ is the number of source examples, $m_T = \beta m$ is the number of target examples, $D = \sqrt{d} / A$, and
\begin{equation}
\nu = \frac{m_T}{m_T + m_S} \left( 1 + \frac{m_S}{\sqrt{D^2(m_S+m_T) - m_S m_T}} \right)
\end{equation}
\begin{equation}
A = \frac{1}{2} \hat{d}_{\mathcal{H} \Delta \mathcal{H}}(\mathcal{U}_S, \mathcal{U}_T) + 4 \sqrt{\frac{2d \log(2m') + \log(\frac{4}{\delta})}{m'}} + \lambda
\end{equation}
\begin{equation}
B = 4 \sqrt{\frac{2d \log(2(m+1)) + 2 \log(\frac{8}{\delta})}{m}}
\end{equation}

\subsection{Discussion}
\subsubsection{Unsupervised}
Equation~\ref{theory1} indicates that if the optimal predictor error $\lambda^*$ on both source and target data is large, then there is no good hypothesis from training on the source domain that will work well on the target domain \cite{ben2010ml,zhao2019learning}. However, as is more common in the application of domain adaptation, if $\lambda^*$ is small, then the bound depends on the source error and the $\mathcal{H} \Delta \mathcal{H}$-divergence \cite{ben2010ml}. The domain-invariant feature learning methods discussed in Section~\ref{domainInvariance} try minimizing these two terms \cite{zhao2019learning}: the source error via a task loss on labeled source data and divergence via a divergence measure such as MMD, with reconstruction, or adversarially. While Section~\ref{results} shows that on many datasets these methods work, there is no guarantee that such adaptation will increase performance (these are upper bounds), as shown by simple counterexamples \cite{zhao2019learning}. It may actually decrease performance if the marginal label distributions differ significantly between source and target \cite{zhao2019learning}.

Equation~\ref{theory2} shows that the target error upper bound alternatively involves the marginal distributions and Equation~\ref{theory3} shows that the lower bound does too. These indicate the importance of aligning the label distributions. If the marginal label distributions are significantly different, then minimizing the source error and divergence between feature representations will actually increase the error \cite{zhao2019learning}. Thus over-training domain-invariant feature learning methods can increase target error, and Zhao et al. \cite{zhao2019learning} experimentally verified this. They found on MNIST, USPS, and SVHN adaptation that during training the target accuracy would initially rise rapidly but would eventually decrease again despite increasing source accuracy, an effect even more apparent with larger differences in the marginal label distributions. It is an open problem as to when the label distributions can be aligned without target labels \cite{zhao2019learning}.

\subsubsection{Semi-Supervised}
Equation~\ref{benLabelBound} indicates that when only source or target data are available, that data should be used (as we might expect). If the source and target are the same, then $\alpha^* = \beta$, which implies a uniform weighting of examples. Given enough target data, source data should not be used at all because it might increase the test-time error. Furthermore, without enough source data using it may also not be worthwhile, i.e., $\alpha^* \approx 0$ \cite{ben2010ml}. In this paper we focus on unsupervised domain adaptation, but these are important considerations if target labels can be obtained. For example, this shows that it may be better to perform semi-supervised adaptation if some labeled target examples are available rather than using the labeled target examples to hyperparameter tune an unsupervised adaptation method.

\section{Applications} \label{applications}
Domain adaptation has been applied in a variety of areas including computer vision, natural language processing, and for time-series data. Using domain adaptation in these various problems can save the human time that would be spent labeling the target data. In some cases such as image semantic segmentation, providing ground truth is very labor intensive. Each pixel-level annotated image in the Cityscapes dataset took on average 1.5 hours to complete \cite{cordts2016cvpr}. In addition, similar methods as described in this paper have been applied to the related problem of domain generalization and some other problems as well.

\subsection{Computer Vision}
Most of the methods surveyed in this paper are for computer vision tasks such as adapting a model trained on synthetic images to real photos (e.g., from synthetic numbers or signs, Table~\ref{comparePerformance1}), stock photos to real photos (e.g., Amazon to DSLR on the Office dataset, Table~\ref{comparePerformance2}), or simple to complex images (e.g., MNIST to SVHN, Table~\ref{comparePerformance1}). Others have been used in robotics for robot grasping \cite{bousmalis2018roboticgrasping}, autonomous navigation \cite{yoo2017domain}, and lifelong learning \cite{wulfmeier2018ieee}, for semantic segmentation \cite{chen2018cvprroad,luo2018taking,lee2018spigan,vu2018advent,huang2018eccv,zou2018eccv,hong2018cvpr,sankaranarayanan2018cvprsemantic,tsai2018cvpr} including when additional information is available from a simulator \cite{lee2018spigan}, in a medical context for chest X-ray segmentation \cite{chen2018seuda}, 3D CT scans to X-ray segmentation \cite{zhang2018tdgan}, MRI to CT scan segmentation \cite{chen2019synergistic}, and MRI segmentation \cite{perone2018unsupervised}, in low resource situations (where there are very few target data points) \cite{hosseiniasl2019augmented}, in situations with different label sets for each domain \cite{sohn2018unsupervised}, for object detection \cite{inoue2018cvpr,chen2018cvpr,hoffman2016fcns}, for person re-identification \cite{ganin2016jmlr,deng2018cvpr,bak2018eccv,wei2018cvpr,zhong2018cvpr,zhong2019cvpr,li2018cvprworkshop}, and for depth estimation \cite{kundu2018cvpr,atapourabarghouei2018cvpr,mahmood2018reverse}.

\subsection{Natural Language Processing}
Domain adaptation has been used in natural language processing such as for sentiment analysis (Table~\ref{compareSentimentPerformance}, \cite{zhang2017aspect,zhao2017principled}), other text classification \cite{liu2017adversarial,zhang2017aspect} including weakly-supervised aspect-transfer from one aspect of a dataset to another \cite{zhang2017aspect}, relation extraction \cite{fu2017domain}, semi-supervised sequence labeling \cite{daume2007acl}, semi-supervised question answering \cite{yang2017semi}, sentence specificity \cite{ko2018domain}, and neural machine translation \cite{chu2018survey,britz2017effective,chen2017cost}.

\subsection{Time Series}
For time-series data, domain adaptation has been used for learning temporal latent relationships in health data across different population age groups \cite{purushotham2017variational}, to perform speech recognition \cite{zhao2017principled,shinohara2016interspeech,hosseiniasl2019augmented}, for predicting driving maneuvers \cite{tonutti2019robust}, anomaly detection \cite{vercruyssen2017transfer}, and inertial tracking \cite{chen2019motiontransformer}. In a method addressing the related problem of domain generalization, time-series radio data was used for sleep-stage classification \cite{zhao2017icml}. Finally, a combination of pre-training and fine-tuning was used to solve another transfer learning problem, where the source datasets have a different label space than the target dataset \cite{fawaz2018tltimeseries}.

\subsection{Domain Generalization} \label{domainGeneralization}
Domain-invariant feature learning approaches similar to those discussed in Section~\ref{domainInvariance} have been used for the related problem of domain generalization, where there are multiple source domains and an unseen target domain \cite{blanchard2011dg,muandet2013domain}. Zhao et al. \cite{zhao2017icml} use an adversarial approach with a domain classifier to learn a model on a dataset collected from a number of people sleeping in various environments that will generalize well to new people and/or new environments (e.g., sleeping in a different room). Ghifary et al. \cite{ghifary2015iccv} use a reconstruction approach with a denoising autoencoder to improve object recognition generalizability, where the ``noise'' is different views (domains) of the data (e.g., rotation, change in size, or variation in lighting) and the autoencoder tries to reconstruct corresponding views of the object in other domains. Carlucci et al. \cite{carlucci2018agnostic} propose an adversarial approach combining domain adaptation and generalization while also doing domain mapping. Akuzawa et al. \cite{akuzawa2019adversarial} note the domain-invariance objective may compete with the discriminative objective and thus develop a method to find the most domain-invariant representation that does not hurt classification performance. Li et al. \cite{li2018eccv} note that previous domain-invariant methods typically assume balanced classes and develop a method to handle changes in class proportions.

\subsection{Other Problems}
Adversarial losses like those used in adversarial domain adaptation methods have also been applied in multiple other settings. Wang et al. \cite{wang2017cvpr} created an adversarial spacial dropout network to add occlusions to images to improve the accuracy of object detection algorithms. They also created an adversarial spatial transformer network to add deformations such as rotations to objects to again increase object detection accuracy. Pinto et al. \cite{pinto2017icra} used adversarial agents to improve a robot's ability to grasp an object via self-supervised learning by employing both shaking and snatching adversaries. Giu et al. \cite{guiteaching} used an adversarial loss to predict and demonstrate (i.e., robot will copy) human motion. Rippel et al. \cite{waveone2017,rippel2018using} used a reconstruction and adversarial loss with an autoencoder for learning higher quality image compression at low bit rates. Sinclair \cite{sinclair2018sounderfeit} applied adversarial loss to clone a physical model for real-time sound synthesis. Adversarial techniques may also be applied to machine learning security, where the goal is to train a classifier robust to adversarial examples \cite{huang2011adversarial,miyato2018virtual}.

\section{Research Directions} \label{researchDirections}
As we have seen, the rapidly-growing body of research focused on unsupervised deep domain adaptation now encompasses many novel methods and components. Here we look at what could be explored in future research to further enhance this existing work.

\subsection{Bi-Directional Adaptation}
The more difficult domain adaptation problems are far from being solved. Tables \ref{comparePerformance1} through \ref{comparePerformanceDatasets} indicate that some domain adaptation problems are harder than others and point to the challenge that more work needs to be focused on these harder problems. While accuracy for SVHN$\rightarrow$MNIST ranges from 70.7\% to 99.3\%, for the reverse case of MNIST$\rightarrow$SVHN, the highest without highly-problem-specific hyperparameter tuning is 81.7\% by Kumar et al. \cite{kumar2018nips} (though tuned on a small amount of labeled target data). This indicates how this reverse problem is much harder \cite{ganin2015icml,french2018iclr}. As a result, few papers offer results for this direction. French et al. \cite{french2018iclr} were able to vastly improve performance up to 97.0\%; however, this required developing a problem-specific unsupervised hyperparameter tuning method. Other methods may similarly benefit from such tuning. Continued work is needed to strengthen general-purpose bi-directional adaptation.

\subsection{Hyperparameter Tuning} \label{hyperparameterTuningFuture}
Some methods such as reverse validation and a problem-specific pixel intensity matching have been applied to hyperparameter tuning without requiring target labels (Section~\ref{hyperparameterTuningExisting}). While the reverse validation method appears promising, it was not used in most of the methods surveyed (only \cite{ganin2016jmlr,pinheiro2018cvpr,pei2018multi}). This may be because of the increase in computation cost \cite{perone2018unsupervised} or problems with the reverse validation accuracy not aligning with test accuracy \cite{bousmalis2016nips}. It is also possible researchers may just be unaware of the method since in the surveyed papers few mention the idea (only \cite{bousmalis2016nips,perone2018unsupervised,ganin2016jmlr,pinheiro2018cvpr,pei2018multi}). Problem-specific methods such as matching pixel intensity between domains as done by French et al. \cite{french2018iclr} are possible given some domain knowledge, but hyperparameter tuning methodologies should be developed that will work across a wider range of problems. This remains an open area of research.

\subsection{Combining Promising Methods} \label{combinePromisingMethods}
French et al. \cite{french2018iclr}, Co-DA \cite{kumar2018nips}, CAN \cite{kang2019contrastive}, AutoDIAL \cite{carlucci2017autodial}, Generate to Adapt \cite{sankaranarayanan2018cvpr}, and WDGRL \cite{shen2018wasserstein} are promising approaches based on Tables~\ref{comparePerformance1} through \ref{compareSentimentPerformance}. French et al. uses a student and teacher network for self-ensembling, Co-DA trains multiple (e.g., two) adaptation networks while requiring diversity and agreement in addition to incorporating virtual adversarial training, CAN alternates between clustering and adaptation through minimizing intra-class discrepancy and maximizing inter-class margin, AutoDIAL adjusts batch normalization layer weights, Generate to Adapt uses an embedding-conditional GAN for adversarial domain adaptation, and WDGRL performs adversarial domain adaptation similar to DANN by using a domain classifier. These are largely independent ideas that if combined may result in additional performance gains.

For instance, the student network in French et al. that accepts either a source or target augmented image could be replaced by the AutoDIAL network to learn how much adaptation to perform at each level of the network. Or to combine with adversarial methods, the student and teacher networks' outputs (or an intermediate layer's outputs, as is being explored by Wang et al. \cite{wang2018domain}) could be fed to a gradient reversal layer followed by a domain classifier, in effect adding an adversarial loss term to the existing two terms used by French et al. Or since French et al. is based upon data augmentation, one might try replacing the existing stochastic data augmentation with a GAN since a GAN can be used for data augmentation (given enough unlabeled training data).

Alternatively, key aspects of other methods could be incorporated. While domain adaptation methods commonly align feature distributions, a different line of research aligns the joint or conditional distribution of the feature and label spaces instead \cite{long2018nips,long2017jmmd,courty2017nips,damodaran2018deepjdot,tang2019discriminative,yu2019transfer,ma2019cvpr,cicek2019iccv}. Researchers found aligning in this manner improves results when handling multi-modal data distributions \cite{long2018nips} or when label proportions differ between domains \cite{courty2017nips}. Other domain adaptation strategies may similarly benefit from aligning the joint or conditional distribution rather than merely the feature distribution.

\subsection{Balancing Classes}
In order to obtain high accuracy on the challenging problem of MNIST$\rightarrow$SVHN, French et al. \cite{french2018iclr} include an additional class-balance term in their loss function, which both improved training stability and helped the network avoid a degenerate local minimum. Though, this term was not required in their other experiments. Clearly, class balancing is an important concern; although, this depends on the dataset being used. Other methods may similarly benefit from balancing classes.

For instance, Hoffman et al. \cite{hoffman2018icml} note that the frequency-weighted intersection over union results in their paper were very close to the target-only model accuracy (an approximate upper bound). Thus, they conclude that domain mapping followed by domain-invariant feature learning is very effective for the common classes in the SYNTHIA dataset (season adaptation on a synthetic driving dataset). It is possible then that additional balancing of classes could help the not-as-common classes to perform better. In addition, data augmentation through occluding parts of the images may improve class balancing as would the adversarial spatial dropout network by Wang et al. \cite{wang2017cvpr} since the two best classes (road and sky) were likely in almost every image.

\subsection{Incorporating Improved Image-to-Image Translation Methods}
Bousmalis et al. \cite{bousmalis2017cvpr} with PixelDA had difficulty applying their method with large domain differences. However, other image-to-image translation methods like XGAN \cite{royer2017xgan} have been developed that may support larger domain shifts. These methods could be extended to domain adaptation directly or also incorporating a semantic consistency loss (as explained in Section~\ref{losses}). This may allow for more substantial differences between domains. Similarly, image-to-image translation methods like StarGAN \cite{choi2018cvpr} have been developed for multiple domains, which could be extended for multi-domain adaptation.

\subsection{Futher Experimental Comparison Between Methods}
As shown in Table~\ref{comparePerformance1}, French et al. \cite{french2018iclr} outperforms all the other methods and Co-DA \cite{kumar2018nips} is quite close behind (with the advantage that it does not require highly-problem-specific tuning on MNIST$\rightarrow$SVHN). In Table~\ref{comparePerformance2}, CAN \cite{kang2019contrastive} outperforms the others followed by Generate to Adapt \cite{sankaranarayanan2018cvpr}. Finally, in Table~\ref{compareSentimentPerformance}, WDGRL \cite{shen2018wasserstein} generally performs the best. However, these methods are not all compared on the same dataset, making a direct comparison difficult. Additional experiments must be performed to see how these methods compare. Similarly, other promising approaches may outperform other methods on some datasets, which could be determined through additional experiments.

These comparisons can be made easier through developing a unified implementation of these various methods. Schneider et al. \cite{schneider2018salad} are developing such an open-source set of implementations of state-of-the-art domain adaptation (and domain generalization) methods. The results provided in individual papers have different hyperparameters, data augmentation, network architectures, etc. that can make direct comparisons challenging. Using a unified implementation of these methods can facilitate more clearly understanding what aspects of a method are responsible for performance gains and also support combining the novel elements from multiple methods.

\subsection{Limitations of Datasets}
Varying amounts of source and target data are available in different situations. The datasets used for comparisons (the image datasets listed in Table~\ref{comparePerformanceDatasets} and the Amazon review dataset) are relatively small when compared with the sizes of datasets commonly in use in deep learning, e.g., ImageNet \cite{5206848,ILSVRC15} (though ImageNet is often used to pretrain adaptation networks). For example, Sankaranarayanan et al. \cite{sankaranarayanan2018cvpr} note how GANs require a lot of training data. This may limit GAN-based methods from being used on too small of source or target datasets. Modifications may need to be developed for such low resource situations, an area explored by Hosseini-Asl et al. \cite{hosseiniasl2019augmented}. Additionally, most domain adaptation datasets are for computer vision. To spur research in other application areas, other datasets could be created.

\subsection{Other Applications} \label{otherApplications}
Other application areas may benefit from performing domain adaptation as have those discussed in Section~\ref{applications}. In particular, only a few methods were applied to time-series data. One time-series application that may benefit from adaptation is activity prediction, e.g., adapting from one type of sensor to another or from one person's data to another's. Some added challenges in this context may be the large differences in feature spaces due to the wide variety of sensors used (e.g., an event stream of fixed motion sensors turning on and off in a smart home vs. sampled motion and location data collected from smart phones or watches) or the difference in labels (e.g., one model may learn a ``walk'' activity while another learns ``exercise'' or may learn ``read'' while another model learns ``school''). Applying domain adaptation in new areas may yield novel methods or components applicable in other areas as well.

\subsection{Other Domain Adaptation Cases}
As mentioned in Section~\ref{methods}, we have surveyed single-source homogeneous unsupervised domain adaptation methods due to this being the most commonly-studied case of domain adaptation. However, exploring other cases is warranted. By utilizing data from multiple source domains and/or multiple target domains, additional gains in performance may be achievable. By handling heterogeneous feature spaces or various other levels of supervision (e.g., semi-supervised learning \cite{saito2019iccv} or weakly-supervised learning \cite{shu2019transferable}), domain adaptation may bring performance gains to other problems as well. Finally, another under-studied case of domain adaptation is partial domain adaptation, where the target domain contains only a subset of the source domain's labels \cite{cao2018cvpr,zhang2018partialda,tang2019discriminative}.

\section{Conclusions}
For supervised learning, deep neural networks are in prevalent use, but these networks require large labeled datasets for training. Unsupervised domain adaptation can be used to adapt deep networks to possibly-smaller datasets that may not even have target labels. Several categories of methods have been developed for this goal: domain-invariant feature learning, domain mapping, normalization statistics-based, and ensemble-based methods. These various methods have some unique and common elements as we have discussed. Additionally, theoretical results provide some insight into empirical observations. Some methods appear very promising, but further research is required for direct comparisons, novel method combinations, improved bi-directional adaptation, and use for novel datasets and applications.

\begin{acks}
This material is based upon work supported by the \grantsponsor{nsf}{National Science Foundation}{https://www.nsf.gov/} under Grant Nos. \grantnum{nsf}{1543656} and \grantnum{nsf}{1734558}.
\end{acks}

\bibliographystyle{ACM-Reference-Format}
\bibliography{bibliography}

\end{document}